\documentclass[letterpaper]{article} % DO NOT CHANGE THIS
\usepackage[preprint]{aaai2027}  % DO NOT CHANGE THIS
\usepackage[hyphens]{url}  % DO NOT CHANGE THIS
\usepackage{graphicx} % DO NOT CHANGE THIS
\usepackage{natbib}  % DO NOT CHANGE THIS AND DO NOT ADD ANY OPTIONS TO IT
\usepackage{caption} % DO NOT CHANGE THIS AND DO NOT ADD ANY OPTIONS TO IT
\usepackage{booktabs}
\usepackage{amsmath}
\usepackage{amssymb}
\usepackage{amsfonts}
\usepackage{mathtools}
\usepackage{paralist}
\usepackage{pifont}% http://ctan.org/pkg/pifont
\usepackage{rotating}
\newcommand{\cmark}{\ding{51}}%
\newcommand{\xmark}{\ding{55}}%

\def\our{ICED}
\title{In-Context Density Estimation for Tabular Data% via Prior-Fitted Energy Networks
}

\author{
    Patryk Marszałek$^{1, 2}$\corresponding,
    Jacek Tabor$^{1}$,
    Marek Śmieja$^{1}$
}
\affiliations{
    \textsuperscript{1}Faculty of Mathematics and Computer Science, Jagiellonian University, Kraków, Poland \\
    \textsuperscript{2}Doctoral School of Exact and Natural Sciences, Jagiellonian University, Kraków, Poland \\
    patryk.marszalek@doctoral.uj.edu.pl, jacek.tabor@uj.edu.pl, marek.smieja@uj.edu.pl
}

\begin{document}

\maketitle

\begin{abstract}
Density estimation underlies many unsupervised tasks on tabular
data such as anomaly detection, out-of-distribution detection, and data
augmentation. Although all these problems reduce to questions about where probability mass lies, they are typically solved individually by fitting a separate model to each dataset, with its
own hyperparameters and tuning budget. We introduce \textbf{\our{}}, an
in-context, energy-based density estimator that removes this per-dataset cost.
\our{} is a transformer-based model pretrained once on a synthetic prior built specifically for
density estimation under an objective
that fits log-density where it is informative and preserves its ordering
elsewhere. In the inference, it
reads a dataset as context and returns an unnormalized log-density for any
query point in a single forward pass, with no fitting, sampling, or
hyperparameter selection. A single frozen \our{} model then drives four tasks usually handled
by four specialized pipelines: density estimation, out-of-distribution
detection, unsupervised anomaly detection, and generative augmentation. Across
all four, it is competitive with the strongest task-specific method, while being
the only approach that needs no retraining, no tuning, and no labels to move
between them. The code is available at \url{https://github.com/gmum/iced}.
\end{abstract}

\section{Introduction}

Estimating probability density is one of the most fundamental problems in
machine learning. A model that accurately captures where probability mass lies
can be used directly for unsupervised anomaly detection, out-of-distribution
(OOD) detection, sample generation, and uncertainty
estimation~\citep{bishop2006pattern}. In this sense density is a central
unsupervised primitive, since many seemingly distinct tasks are, at their core,
the same question about where the data concentrate. Yet despite these shared
probabilistic foundations, the corresponding problems are typically treated in
isolation, each addressed by its own specialized method. And each such method must be fit anew on every dataset, tuned,
and often re-engineered for the task at hand.

This per-dataset, per-task optimization paradigm stands in sharp contrast to
recent progress in foundation models. In natural language processing and computer
vision, task-specific optimization has increasingly given way to large models
that acquire general inference capabilities during pretraining and adapt to new
tasks purely through conditioning~\citep{brown2020language}. A similar shift has
recently reached tabular supervised learning through Prior-Data Fitted Networks
(PFNs), which learn a prediction algorithm over synthetic tasks and solve unseen
classification or regression problems in a single forward pass, without
retraining or hyperparameter
optimization~\citep{muller2022transformers,hollmann2023tabpfn,hollmann2025tabpfnv2}.
This line of work suggests that \emph{learning the algorithm}, rather than
fitting a model from scratch for every dataset, is a powerful alternative for
tabular machine learning.

\begin{figure}[t]
  \centering
  \includegraphics[width=\columnwidth]{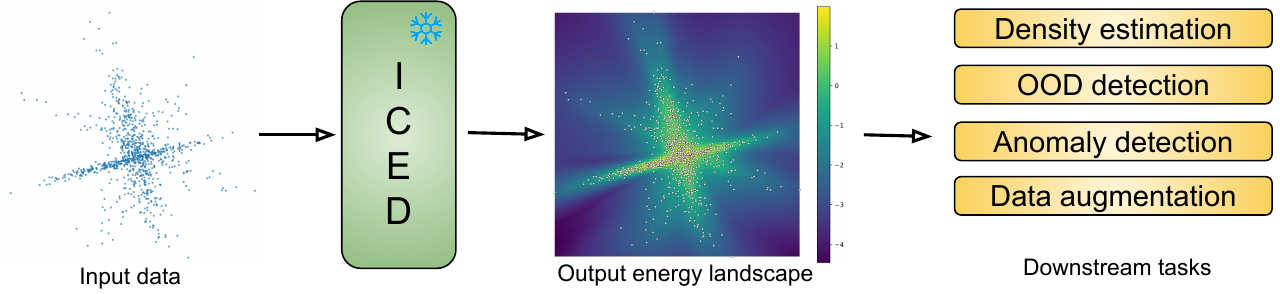}
  \caption{\textbf{\our{} turns density estimation into a single reusable model.}
  Pretrained once, a \emph{frozen} \our{} serves density estimation, out-of-distribution detection,
  unsupervised anomaly detection, and data augmentation, with no retraining,
  tuning, or labels.}
  \label{fig:teaser}
\end{figure}

Surprisingly, density estimation has not undergone this transition. The
established estimators, including kernel density
estimation~\citep{scott2015multivariate,silverman1986density}, Gaussian
mixtures, normalizing flows~\citep{papamakarios2021normalizing,dinh2017density},
and energy-based models~\citep{lecun2006tutorial}, all remain firmly in the
per-dataset regime, learning a fresh model for each new table. In principle, an
in-context predictor such as TabPFN can already produce densities, by
autoregressively factorizing the joint over features one dimension at a
time~\citep{hollmann2023tabpfn}. However, this procedure is slow, scales poorly with
dimension, and can be unreliable, since density is recovered only as a
by-product of a model trained for supervised prediction. More recent PFN-based
models tackle individual unsupervised problems, but do so
discriminatively~\citep{zhao2026tabclustpfn,marszalek2025zeus,tactic2026}: each
amortizes a single task-specific output rather than the density that underlies
them. As a result, practitioners still maintain a separate pipeline for each
unsupervised task, despite many of them resting on the same notion of
probability density.

In this work, we introduce \textbf{\our{}}, an \textbf{I}n-\textbf{C}ontext
\textbf{E}nergy-based \textbf{D}ensity estimator that removes this optimization
loop entirely, see Figure~\ref{fig:teaser}. Instead of learning a density model for every dataset, \our{}
learns the density-estimation algorithm itself. Following the PFN paradigm, it
is a transformer pretrained once on a carefully designed synthetic prior
spanning multimodal, heavy-tailed, nonlinear, and mixed-type tabular
distributions, for which exact log-densities are available by construction. At
inference, the pretrained transformer receives an unseen dataset as context
together with arbitrary query points and predicts their energies in a single
forward pass (Figure~\ref{fig:overview}), requiring neither gradient updates,
sampling, nor hyperparameter tuning.

Representing density through an energy (an unnormalized log-density) is
central to this generality. Most downstream applications depend only on
\emph{relative} density rather than a normalized likelihood, so an energy
discards the one quantity that is intractable to compute, the partition
function, while retaining everything these tasks
require~\citep{lecun2006tutorial}. A single energy landscape therefore supports
many tasks at once. The same frozen \our{} model ranks samples by density,
flags anomalies and OOD examples through its low-energy regions, and generates
new in-distribution samples by following the energy gradient with Langevin
dynamics~\citep{welling2011bayesian}, all without task-specific retraining or
architectural change.

Our contributions are summarized as follows:
\begin{compactitem}
  \item We introduce, to our knowledge, the first in-context foundation model
        for tabular density estimation, extending the PFN paradigm from
        supervised prediction to unsupervised probabilistic modeling.
  \item We design a synthetic pretraining prior and a training objective
        tailored to density estimation, combining exact synthetic log-densities,
        broad distributional diversity, and a hybrid regression-ranking loss
        that concentrates model capacity on the informative regions of the
        density landscape.
  \item We demonstrate that a single frozen pretrained model is competitive with
        specialized methods across four traditionally separate
        tasks (density estimation, OOD detection, anomaly detection, and data
        augmentation) while requiring no retraining, no hyperparameter search,
        and no task-specific adaptation.
\end{compactitem}

\section{Related Work}
\label{sec:related}

\paragraph{Tabular foundation models and in-context learning.}
Prior-fitted networks (PFNs) reframe Bayesian inference as supervised learning:
a transformer is pretrained on datasets sampled from a chosen prior so that one
forward pass approximates the posterior predictive
distribution~\citep{muller2022transformers}. TabPFN brought this to tabular
classification, pretraining on a structural-causal-model prior and solving
small tasks in a single pass with no per-dataset
training~\citep{hollmann2023tabpfn}; later work scaled it to larger tables and
to regression~\citep{hollmann2025tabpfnv2,grinsztajn2025tabpfn25}, built
fixed-dimensional row embeddings for very large tables~\citep{qu2025tabicl}, and
combined in-context retrieval with self-supervised pretraining on real
tables~\citep{ma2024tabdpt}. 
% \our{} shares the amortised, single-forward-pass
% philosophy of this family, but departs from its defining assumption: these
% models consume a \emph{labelled} context and emit a \emph{label}. \our{} consumes
% an unlabelled context and emits a density over the feature space---a generative
% quantity, not a discriminative one.

% \paragraph{Amortising unsupervised tasks.}
A newer line carries the PFN recipe into unsupervised problems, and is the
closest precedent for our work. TabClustPFN amortizes clustering, inferring
cluster assignments and their number in one pass~\citep{zhao2026tabclustpfn};
ZEUS produces zero-shot embeddings on which off-the-shelf clusterers
operate~\citep{marszalek2025zeus}; TACTIC casts anomaly detection as in-context
inference, returning calibrated outlier probabilities~\citep{tactic2026}. These
methods establish that prior-fitting generalizes well beyond classification.
Crucially, however, each amortizes a single task-specific output, such as a
partition, an embedding, or an anomaly score, rather than the density that
underlies them.
% Our
% thesis is that these outputs are downstream of a common quantity, density, and
% that amortising density itself yields all of them at once. \our{} therefore does
% not compete with these methods task-for-task so much as subsume the quantity
% they each specialise; a single \our{} model spans the OOD, anomaly, and
% generative settings that would otherwise require three of them.

\paragraph{Density estimation.}
Classical density estimators trade flexibility for per-dataset fitting. KDE places a
kernel at each sample but is bandwidth-sensitive and degrades sharply in
dimension~\citep{silverman1986density,scott2015multivariate}. Gaussian mixtures
require choosing a component count, e.g.\ by BIC~\citep{schwarz1978estimating}.
Deep likelihood models are more expressive: normalizing flows give exact
likelihoods through invertible maps~\citep{dinh2017density,papamakarios2017masked,kingma2018glow,durkan2019neural,papamakarios2021normalizing},
autoregressive models factorize the joint into ordered
conditionals~\citep{germain2015made,vandenoord2016pixel}, and variational
autoencoders optimize a likelihood lower bound~\citep{kingma2014auto}. However, each of these models
is trained anew per dataset, which requires separate optimization and parameter selection. 
% \our{} keeps exact-likelihood synthetic supervision (its prior is built
% from these very models) yet removes the per-dataset training they all require.
% \paragraph{Energy-based models.}

Energy-based models (EBMs) learn an unnormalized energy and sidestep the
partition function~\citep{lecun2006tutorial}. Energies can be fit by score
matching~\citep{hyvarinen2005estimation} and sampled by Langevin
dynamics~\citep{song2019generative,welling2011bayesian}. Closest to our setting,
TabEBM constructs a distinct class-specific EBM for tabular data augmentation by
turning a frozen in-context classifier into an energy through a surrogate binary
task, and samples from it with SGLD~\citep{margeloiu2024tabebm}. \our{} shares
this energy-based, in-context, training-free lineage, but differs in what the
energy represents. TabEBM builds an energy indirectly, from
classifier logits on a surrogate task, and targets class-conditional generation.
\our{} instead predicts a density $\log p(x)$ directly, from
a single model that needs no surrogate construction, which
is what lets the same energies serve ranking, detection, and generation alike.
%; JEM reinterprets a classifier as an EBM of the joint~\citep{grathwohl2020your}. 
% Closest to us,
% TabEBM fits a \emph{distinct} class-specific EBM for tabular augmentation and
% samples it with SGLD~\citep{margeloiu2024tabebm}. We inherit TabEBM's
% energy-based, tabular lineage and its SGLD sampler, but invert the cost
% structure: where TabEBM fits a new energy model per dataset and per class, \our{}
% carries one frozen energy model across all datasets, and targets unsupervised
% joint density rather than class-conditional sampling alone.

% =====================================================================
\section{Method}
\label{sec:method}

We present \textbf{\our{}}, an in-context energy-based density estimator for
tabular data, pretrained once on a synthetic prior and applied to any dataset
in a single forward pass. We first formalize in-context density estimation and
the energy formulation we adopt (Section~\ref{sec:method-setup}), then describe
the two ingredients that determine what the model learns: the synthetic prior it
is trained on (Section~\ref{sec:method-prior}) and the training objective
(Section~\ref{sec:method-loss}). We then specify the model architecture
(Section~\ref{sec:method-arch}) and describe how a single frozen model serves all
downstream tasks at inference (Section~\ref{sec:method-inference}).

\subsection{In-context density estimation perspective}
\label{sec:method-setup}

\begin{figure*}[t]
  \centering
  \includegraphics[width=\textwidth]{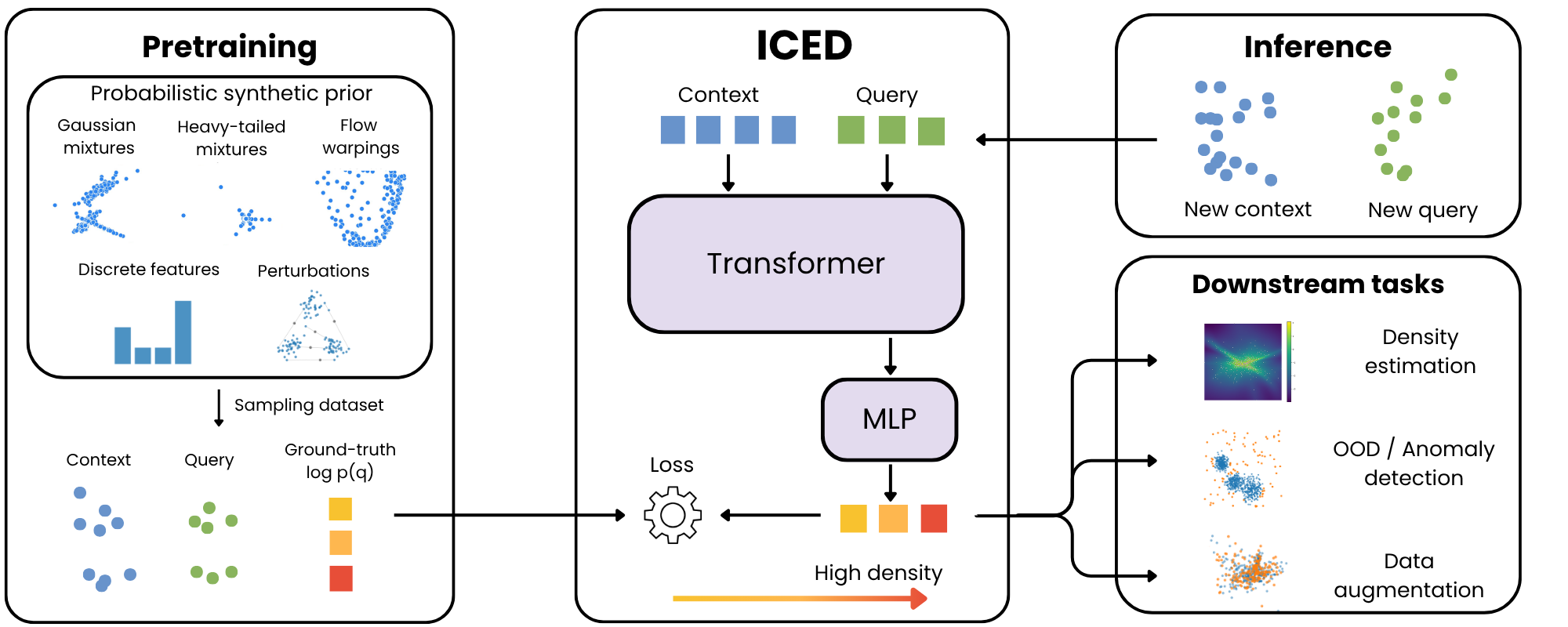}
  \caption{\textbf{Overview of \our{}.} A transformer is pretrained \emph{once} on
  a synthetic prior with exact log-densities (Gaussian and heavy-tailed mixtures,
  flow-warped distributions, mixed discrete/continuous features). At inference, a
  \emph{single frozen} model reads a dataset as context $D$ together with query
  points $Q$ and returns an energy for each query
  in a single forward pass. The same energies then drive four tasks that are usually
  handled by separate, per-dataset pipelines.}
  \label{fig:overview}
\end{figure*}

% \paragraph{Problem setup.}
Let $D = \{x_i\}_{i=1}^{n} \subset \mathbb{R}^d$ be a tabular dataset drawn
i.i.d.\ from an unknown distribution $p$, and let $Q = \{q_j\}_{j=1}^{m} \subset \mathbb{R}^d$ be a
set of query points. Our goal is to estimate the (log-)density of $p$ at each
query. Conventional estimators address this by \emph{fitting} a model to
$D$ (selecting a bandwidth, a number of mixture components, or the weights of a
neural network) and then evaluating it on $Q$, so that every dataset incurs a
fresh fitting procedure.

% \paragraph{The prior-fitting perspective.}
Following the prior-fitted network (PFN)
paradigm~\citep{muller2022transformers,hollmann2023tabpfn}, we instead learn a
\emph{single} model that performs density estimation in context. Rather than
fitting parameters per dataset, we train one network $f_\theta$ on a large
collection of synthetic datasets sampled from a prior $p(\mathcal{D})$ over
data-generating processes. During pretraining, each instance provides a dataset
$D$, query points $Q$, and their ground-truth log-densities; the network learns
to predict these from $D$ and $Q$ alone. Formally, it is trained to approximate
\begin{equation}
  f_\theta(q \mid D) \;\approx\; \mathbb{E}_{p \sim p(\mathcal{D})}
  \big[\, \log p(q) \;\big|\; D \,\big],
\end{equation}
so that, at inference, density estimation on a previously unseen dataset reduces
to a single forward pass: the parameters $\theta$ are shared across all datasets,
and the dependence on $D$ enters only through in-context conditioning. No
gradient steps and no hyperparameter selection are performed at test time, see Figure~\ref{fig:overview} for the illustration.

% \paragraph{Energy rather than likelihood.}
\our{} estimates an \emph{energy}, i.e. an unnormalized log-density, rather than a
normalized one. Concretely, its output approximates $\log p(q)$ only up to a
dataset-dependent additive constant $c_D$ that absorbs the log-partition
function\footnote{Note that we define energy with the opposite sign to the usual EBM
convention, so that higher energy corresponds to higher density.}~\citep{lecun2006tutorial},
\begin{equation}
  f_\theta(q \mid D) \;\approx\; \log p(q) + c_D .
\end{equation}
This choice is deliberate. The tasks that consume tabular densities, such as anomaly
and OOD scoring, and gradient-based generation, are invariant to
$c_D$, depending only on relative density between points. Modeling energy
therefore removes the one quantity that is intractable to compute while
retaining everything the downstream tasks require: higher energy denotes higher
density, so a query in a dense region of $D$ receives a high value and an outlier
a low one.

\subsection{Synthetic prior}
\label{sec:method-prior}

A prior-fitted network is only as good as the prior it is trained on. %: the prior \emph{is} the inductive bias. 
\our{} is trained
entirely on synthetic data, where each training instance is a triple
$(D, Q, \{y_j\}_j)$ with ground-truth log-densities $y_j = \log p(q_j)$ available
by construction. We design the prior to span the distributional structures
encountered in real tabular data while keeping exact
log-densities tractable. Each instance is generated by sampling a base mixture,
optionally warping it through an invertible flow, optionally appending discrete
features, and perturbing a fraction of the points; we describe each in turn.

\paragraph{Base mixtures.}
With equal probability we sample either a Gaussian mixture or a heavy-tailed
mixture of $K \in [2, 20]$ components. Gaussian components use random means and
random positive-definite covariances $\Sigma_k = A_k A_k^\top$, with exact
log-densities from the closed-form mixture log-likelihood (computed stably via
Cholesky factorization). Heavy-tailed mixtures draw each component from a
Student-$t$ ($\nu=3$), Laplace, or Cauchy base distribution under a random
affine map $x = A_k z + b_k$; the change-of-variables rule gives the exact
log-density 
$$
\log p(x) = \log p_0\!\big(A_k^{-1}(x - b_k)\big) - \log|\det
A_k|. 
$$
Including heavy tails teaches the model that density can decay at very
different rates, which is essential for robust anomaly scoring.

\paragraph{Nonlinear warping via flows.}
Real tabular distributions are rarely mixtures of simple unimodal components.
To inject nonlinear geometry while preserving an exact likelihood, we
optionally compose the base mixture with an invertible normalizing
flow~\citep{papamakarios2021normalizing}, applying the change-of-variables
formula to propagate log-densities. We draw from two flow families: the standard
affine coupling flow (RealNVP)~\citep{dinh2017density} and a piecewise flow that
we introduce in this work (see Appendix~\ref{app:flows} for its detailed formulation). The flow
warps both context and query points, yielding distributions with curved,
multimodal support far richer than the base mixtures alone.

\paragraph{Discrete features.}
To reflect the mixed-type nature of tabular data, with probability
$p_{\text{cat}}$ we append one to five independent categorical features. Each
takes $C_j \in \{2,\dots,5\}$ values with class probabilities drawn from a
Dirichlet prior, contributing additively to the joint
log-density. Categorical columns are one-hot encoded and interleaved with the
continuous columns in a random order, preventing the model from exploiting any
fixed positional layout.

\paragraph{Perturbation for a robust landscape.}
A density estimator that has only ever seen exact samples can assign
arbitrary energies just off the data manifold. To shape the learned landscape in
these out-of-distribution regions, we perturb a random $10\%$ of the coordinates of a
fraction of the query points before they are scored, using one of Gaussian
noise (with dimension-adaptive scale), Mixup with a context point, or CutMix
feature swaps (see Appendix~\ref{app:augmentation-details} for details). This populates the region around the manifold with
partially-in-distribution queries and regularizes the energy surface where the
downstream tasks operate.

\paragraph{Target normalization.}
Because we model energy up to an additive constant, only the \emph{shape} of
the log-density matters, not its scale. We therefore normalize the per-instance
targets to a common range using their empirical $10$th and $90$th percentiles
as anchors, $\tilde{y}_j = 2\,(y_j - \hat q_{0.1})/(\hat q_{0.9} - \hat q_{0.1})
- 1$. This stabilises training across instances whose raw log-densities span
wildly different magnitudes.

\subsection{Training objective}
\label{sec:method-loss}

Given an instance, the model predicts energies $\hat y_j = f_\theta(q_j \mid
D)$ and is trained against the normalized targets $\tilde y_j$. Two properties
of density estimation shape our loss. First, we optimize 
the accuracy of estimation
in the high-density regions where the data actually lie. The precise depth of the
tails is dominated by outliers, hard to fit, and rarely used, so we do not spend
capacity on it. Second, several
downstream tasks consume density only through the ordering it induces
over points. We combine two terms accordingly,
\begin{equation}
  \mathcal{L} = \mathcal{L}_{\text{reg}} + \mathcal{L}_{\text{rank}}.
\end{equation}

% \paragraph{Thresholded regression.}
For queries whose target lies in the in-distribution range we use a squared
error; for low-density queries we ask only that their predicted energy stay
below the range, not that it match an exact value:
\begin{equation}
  \mathcal{L}_{\text{reg}} =
  \underbrace{\frac{1}{|\mathcal{I}|}\sum_{j \in \mathcal{I}}
    (\hat y_j - \tilde y_j)^2}_{\text{fit where density is high}}
  \;+\;
  \underbrace{\frac{1}{m}\sum_{j \notin \mathcal{I}}
    \phi(\hat y_j)}_{\text{bound the OOD tail}},
\end{equation}
where $m$ is the number of query points, $\mathcal{I} = \{ j : \tilde y_j \in
[\tau_{\text{lo}}, \tau_{\text{hi}}] \}$ are the in-range queries, and $\phi$ is
a one-sided penalty that activates only when an out-of-range query is predicted
inside the range. This
keeps model capacity focused on the informative, high-density region instead of
chasing the precise depth of every outlier.

% \paragraph{Pairwise ranking.}
To align the model directly with order-based use, we penalise every pair whose
predicted order disagrees with the ground truth,
\begin{equation}
  \mathcal{L}_{\text{rank}} =
  \frac{1}{|\mathcal{P}|}
  \sum_{(j,k) \in \mathcal{P}}
  \ell\!\big(\hat y_j - \hat y_k\big),
\end{equation}
where $\mathcal{P} = \{ (j,k) : \tilde y_j > \tilde y_k \}$ is the set of query
pairs ordered by their target density, and $\ell$ is a margin-based pairwise
loss; we use the logistic form
$\ell(s) = \log(1 + e^{-s})$ in all experiments. This term improves rank metrics
(Kendall's $\tau$, pairwise accuracy) even when absolute energies are
imperfect, which is precisely what anomaly and OOD scoring rely on.

Both $\tau_{\text{lo}}, \tau_{\text{hi}}$ and the loss form are fixed once
during pretraining; crucially, they are properties of \emph{training}, not
knobs exposed at inference. The model is optimized with AdamW under a cosine
schedule, with gradient accumulation and clipping, on a stream of synthetic
instances generated on the fly. During pretraining, the context length varies from 200 to 2000 samples, while the query length is fixed at 256 samples. Furthermore, the pretraining datasets contain a uniformly distributed number of features, ranging from 2 to 50.

\subsection{Model architecture}
\label{sec:method-arch}

\our{} instantiates $f_\theta$ as a transformer that processes context and query
points jointly as a set, in the spirit of TabPFN~\citep{hollmann2023tabpfn} and
its unsupervised successors~\citep{marszalek2025zeus}. Each point is embedded
into a token by the TabICLv2 feature extractor~\citep{qu2026tabiclv2}, which handles
inputs of varying dimensionality. At the beginning of the pretraining process, the feature extractor is initialized with weights from a pretrained TabICLv2 model. It is then fine-tuned jointly with the rest of the model throughout pretraining.
%, including tables whose feature count exceeds the model's nominal width.
%If data dimensionality exceeds $d_{\max}$, we repeatedly subsample features, which forms an ensemble whose predicted probabilities are then averaged, in line with standard PFN practice.
The context and query tokens are processed together by a stack of transformer
encoder layers under an attention mask in which query tokens attend to the full
context while context tokens attend only among themselves. This makes
$f_\theta(q_j \mid D)$ a function of the entire context set, keeps the context
representation independent of the queries, and renders the model
permutation-invariant in the data points. A small multilayer-perceptron head
maps each query token to a scalar energy, so that all $m$ estimates
$\{f_\theta(q_j \mid D)\}_{j=1}^{m}$ are produced in one forward pass. Full
architectural specifications are deferred to Appendix~\ref{app:architecture}.

\subsection{Inference: one model, many tasks}
\label{sec:method-inference}

At test time the user supplies a context dataset $D$ and queries $Q$. Both are
scaled using statistics from $D$, and the
frozen model returns the energies $\{f_\theta(q_j \mid D)\}$ in a single forward
pass. The \emph{same} model and the \emph{same} energies then serve every
downstream task, differing only in how the energy is read:

\begin{compactenum}
  \item \textbf{Density estimation \& ranking.} The energies are used directly
        as log-density proxies, ordering queries from most to least typical.
  \item \textbf{Anomaly / OOD detection.} The negated energy is an outlier
        score: points in low-density regions of $D$ are flagged as anomalous or
        out-of-distribution.
  \item \textbf{Data augmentation.} Treating $f_\theta(\cdot \mid D)$ as an
        energy surface, we draw new in-distribution samples by following its
        gradient with Stochastic Gradient Langevin Dynamics
        (SGLD)~\citep{welling2011bayesian}, conditioning on a class-specific
        context as in TabEBM~\citep{margeloiu2024tabebm}. 
\end{compactenum}
% The practical advantage of casting density estimation as an in-context,
% energy-based problem is that a single pretrained model replaces a family of
% task- and dataset-specific estimators, with no retraining and no
% hyperparameter search.

\section{Experiments}
\label{sec:experiments}

Our experiments demonstrate that one frozen,
prior-fitted density model can stand in for the specialized pipelines normally
built for each individual task. We first establish that \our{} estimates
density faithfully on held-out synthetic distributions
(Section~\ref{sec:exp-density}), a controlled check with known ground truth.
We then take that \emph{same checkpoint, unchanged}, and ask it to perform three
downstream tasks it was never explicitly trained for: out-of-distribution
detection (Section~\ref{sec:exp-ood}), unsupervised anomaly detection
(Section~\ref{sec:exp-anomaly}), and data augmentation
(Section~\ref{sec:exp-aug}). An ablation isolates the contribution of the prior
and the loss (Section~\ref{sec:exp-ablation}). 
% The experimental design is
% itself the central message: across every task the only thing that changes is the
% context dataset handed to the model.

% =====================================================================

\subsection{Density estimation on synthetic data}
\label{sec:exp-density}

% \paragraph{Setup.}
In this proof-of-concept experiment, we ask whether \our{}, pretrained on
the synthetic prior of Section~\ref{sec:method}, estimates $\log p(x)$
correctly on \emph{held-out} data drawn from the same prior but never seen
during pretraining. We group test distributions by their generative class,
following the structure of the prior: Gaussian mixtures, heavy-tailed
mixtures, and mixtures pushed through random invertible flows, with and
without discrete coordinates. For each class we sample 50 datasets, supply
each as context, and score a held-out set of queries whose ground-truth
log-densities are known by construction.

% \paragraph{Baselines.}

% \paragraph{Metrics.}
Because the downstream value of a density estimate lies in the ordering it
induces over points, and because \our{} models an unnormalized energy, we
evaluate \emph{rank} agreement between predicted and ground-truth
log-densities rather than absolute likelihood. We report Kendall's $\tau$ and
a pairwise ordering accuracy
$(C + \tfrac{1}{2}T)/\binom{n}{2}$,
where $C$ and $T$ are the numbers of concordant and tied query pairs,
respectively\footnote{a pair is concordant when the predicted and reference
log-density differences share the same sign}. Pairwise accuracy is reported in
Table~\ref{tab:density-pa}; Kendall's $\tau$, which yields the same ranking of
methods, is deferred to Appendix~\ref{app:density-kendall}. Each reported
value is averaged over the datasets sampled from the corresponding class.

We compare against three families: (i) classical estimators: kernel density
estimation (KDE)~\citep{scott2015multivariate} and Gaussian mixtures with
BIC-selected components (GMM-BIC), (ii) deep density estimators: masked
autoregressive flows (MAF)~\citep{papamakarios2017masked},
RealNVP~\citep{dinh2017density}, and a variational
autoencoder~\citep{kingma2014auto}, (iii) tabular foundation models: autoregressive
TabPFN~\citep{hollmann2023tabpfn}, TabICL~\citep{qu2025tabicl}, and
TabEBM~\citep{margeloiu2024tabebm}. The classical and deep estimators are fit
anew on each dataset, and most also require their own hyperparameter search; the
tabular foundation models are frozen like \our{} but recover density only
indirectly, autoregressively or through a surrogate task, rather than as a
direct output.

\begin{table*}[t]
\centering
\small
\caption{Density estimation on held-out synthetic data. We report pairwise ordering accuracy (higher is better; chance
$=0.5$). \our{} and the in-context baselines (TabPFN, TabICL, TabEBM) use a
single frozen model across all classes, while the classical and deep estimators are fit per dataset. \textbf{Bold} is best,
\underline{underline} second best.}
\label{tab:density-pa}
\begin{tabular}{lcc|ccccccccc}
\toprule
 \multicolumn{3}{c|}{Evaluation datasets} & \multicolumn{9}{c}{Methods} \\
Base class & Cat. & Aug. & KDE & GMM-BIC & MAF & RealNVP & VAE & TabPFN & TabICL & TabEBM & \our{} \\
\midrule
Gaussian mix. & \cmark & \xmark & 0.7355 & \underline{0.7730} & 0.7251 & 0.7339 & 0.7355 & 0.7486 & 0.7671 & 0.5885 & \textbf{0.7824} \\
Heavy-tailed mix. & \cmark & \xmark  & 0.6838 & \underline{0.7241} & 0.6794 & 0.6692 & 0.6500 & 0.7025 & 0.6416 & 0.6006 & \textbf{0.7462} \\
% Mix. w. cat. & 0.6915 & \underline{0.7211} & 0.6824 & 0.6754 & 0.6536 & 0.6844 & 0.6113 & 0.5983 & \textbf{0.7306} \\
% Mix. flow (cont) & 0.7229 & 0.7441 & 0.7163 & 0.7115 & 0.6874 & \underline{0.7464} & 0.7393 & 0.6022 & \textbf{0.7671} \\
Mix. with flows & \cmark & \xmark & 0.6975 & 0.7092 & 0.6985 & 0.6974 & 0.6856 & \underline{0.7153} & 0.7030 & 0.5575 & \textbf{0.7541} \\
All datasets & \xmark & \cmark & 0.6922 & \underline{0.7379} & 0.6801 & 0.6867 & 0.6593 & 0.7066 & 0.6777 & 0.5769 & \textbf{0.7381} \\
All datasets & \cmark & \cmark & 0.6658 & \underline{0.7095} & 0.6621 & 0.6467 & 0.6308 & 0.6782 & 0.6288 & 0.5826 & \textbf{0.7209} \\
% \midrule
% Mean & 0.6984 & \underline{0.7313} & 0.6920 & 0.6887 & 0.6717 & 0.7117 & 0.6812 & 0.5866 & \textbf{0.7485} \\
\bottomrule
\end{tabular}
\end{table*}

% \paragraph{Results.}
The pattern to look for in Table~\ref{tab:density-pa} is not a
single winner but a \emph{robustness gap}: \our{} is the best or second-best
estimator in every distribution class, while the baselines are each strong only
where their assumptions hold. KDE and GMM-BIC, despite being fit directly to the
test data, are competitive on the Gaussian classes and erode on heavy-tailed and
flow-warped ones. The single-dataset flows and VAE track \our{} loosely but never lead. Tabular foundation models are inferior to competitive methods.
% Most tellingly, the autoregressive TabPFN and the VAE---models built for a
% different objective---produce near-inverted orderings on several classes,
% exactly the failure one expects when density is read off a model that was never
% asked to represent it. 
\our{}, alone among the methods, is never fit to the test
data and yet never far from the top. %This is the controlled-setting version of
% the paper's thesis: a prior broad enough to cover these families lets a single
% amortized model recover their densities without seeing them.

\begin{figure}[t]
    \centering
    \includegraphics[width=\columnwidth]{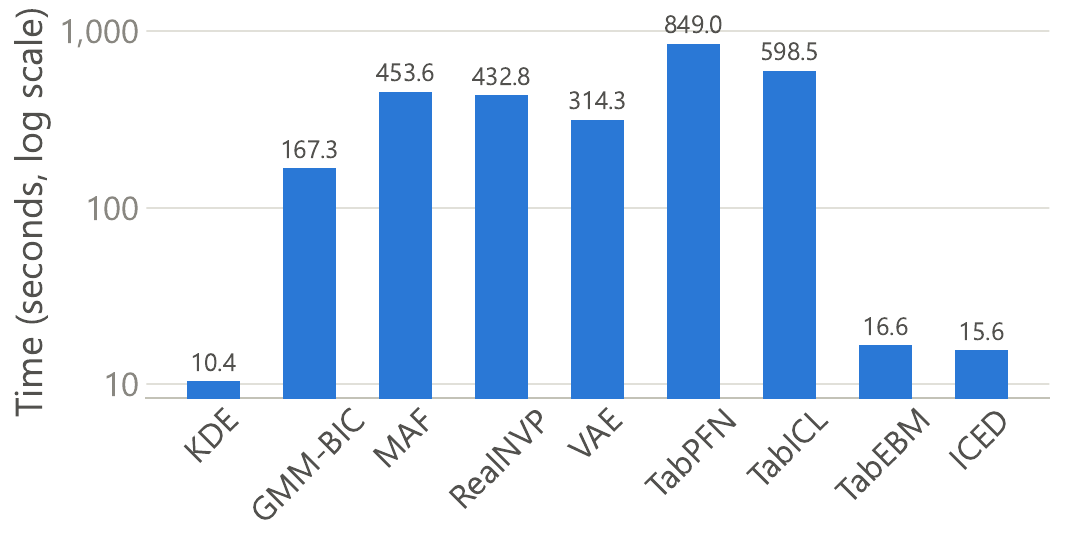}
    \caption{Density-estimation runtime by method (lower is better). \our{}
    performs a single forward pass and is among the fastest methods, whereas the
    per-dataset deep estimators and the autoregressive in-context models (TabPFN,
    TabICL) are substantially slower.}
    \label{fig:times}
\end{figure}

% \paragraph{Inference speed.}
Because \our{} estimates density in a single forward pass, it avoids the
per-dataset fitting loop that the classical and deep baselines pay at test time.
Figure~\ref{fig:times} compares the time to produce density estimates. \our{}
is among the fastest methods, second only to KDE and on par
with TabEBM, while the per-dataset deep estimators are an order of
magnitude slower and the
autoregressive in-context models are slower still, since they recover density one feature dimension at a time. \our{} is
thus roughly $50\times$ faster than TabPFN while producing a joint density
directly, which makes a single frozen model practical to apply across many
datasets, not just accurate.

% =====================================================================
\subsection{Out-of-distribution detection}
\label{sec:exp-ood}

% \paragraph{Setup.}
The first transfer test asks whether \our{}'s energies flag points that fall
outside a reference distribution. Using the ADBench
suite~\citep{han2022adbench}, we give \our{} a context of in-distribution points
only and a query set mixing in- and out-of-distribution points; the negated
energy is the OOD score. We report AUROC averaged over ADBench datasets and five random seeds, and
compare against every ADBench detector together with the density estimators of
Section~\ref{sec:exp-density} fit on the in-distribution context.
% , pitting \our{}
% against both purpose-built detectors and general-purpose density models at once.

\begin{figure}[t]
    \centering
    \includegraphics[width=\columnwidth]{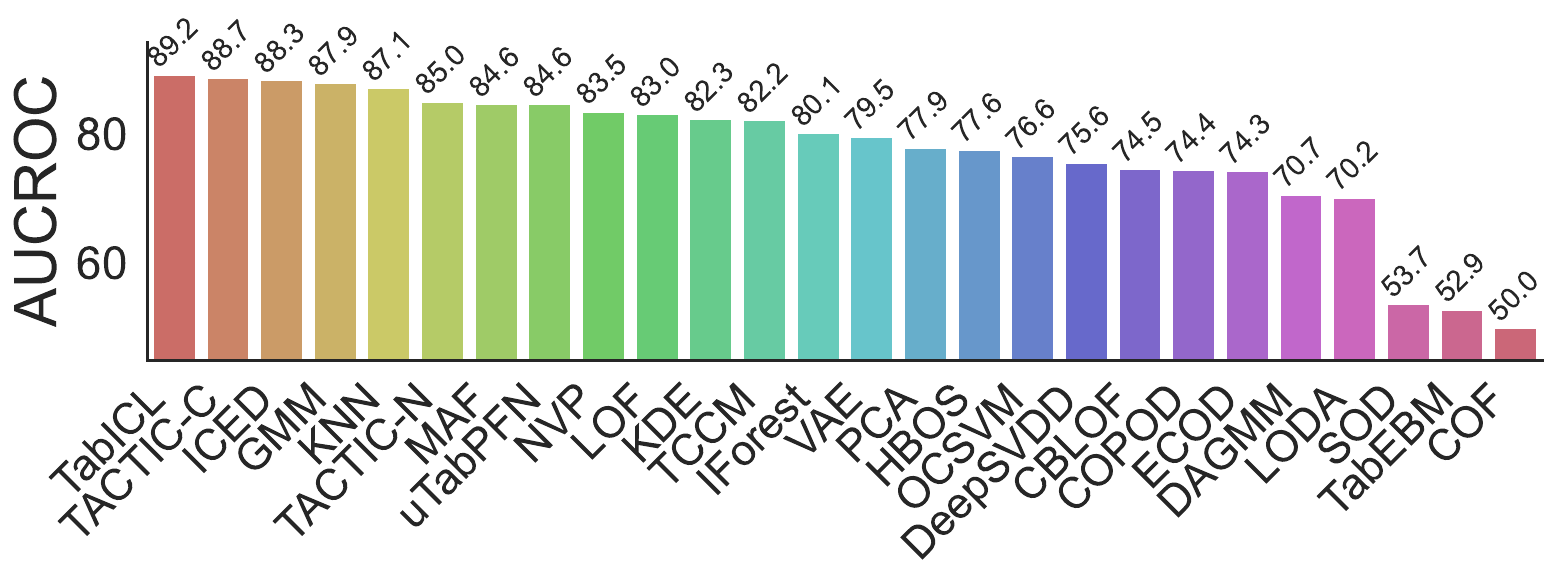}
    \caption{Out-of-distribution detection on ADBench. Mean AUROC over all datasets (higher is better).}
    \label{fig:an-auroc}
\end{figure}

% \paragraph{Results.}
Figure~\ref{fig:an-auroc} presents the mean AUROC. \our{} obtains
$88.3$, placing
third among all $26$ methods. It is within one point of the best, behind only
TabICL ($89.2$) and TACTIC-C ($88.7$) and well ahead of TabPFN ($84.6$) and TabEBM ($52.9$). A clean
reference set rewards methods that capture the global density, so several
classical detectors are close behind, notably GMM ($87.9$) and KNN ($87.1$).
% The detail worth carrying to the
% next section is that the OOD leaders, TabICL, TACTIC-C, and GMM, are not the
% methods that stay on top once the context is contaminated
% (Section~\ref{sec:exp-anomaly}).

% =====================================================================
\subsection{Unsupervised anomaly detection}
\label{sec:exp-anomaly}

% \paragraph{Setup.}
Here the context is contaminated rather than clean: it contains both normal
points and a small, \emph{unlabeled} fraction of anomalies, as specified by
ADBench~\citep{han2022adbench}. 
% Since anomalies are by definition rare, a
% faithful density places them in its low-energy tail, so \our{} can score every
% context point by its own estimated density (no labels, no separate training
% phase). 
As before, we report AUROC over all datasets and five random seeds, with the
same datasets and baselines as in Section~\ref{sec:exp-ood}.

\begin{figure}[t]
    \centering
    \includegraphics[width=\columnwidth]{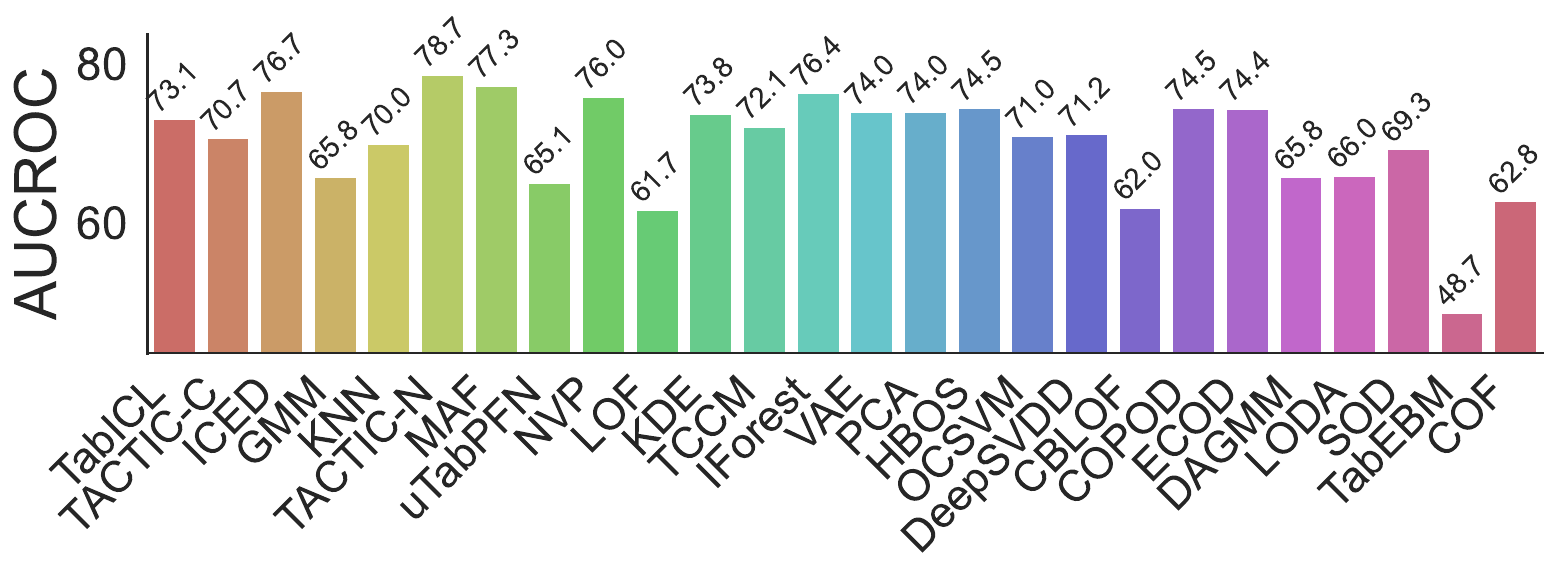}
    \caption{Unsupervised anomaly detection on ADBench. Mean AUROC over all
datasets (higher is better).}
    \label{fig:ood-auroc}
\end{figure}

% \paragraph{Results.}
Under contamination (Figure~\ref{fig:ood-auroc}), \our{} reaches $76.7$ mean
AUROC, again third
overall, behind TACTIC-N ($78.7$) and MAF ($77.3$). The comparison with the
previous section shows that the methods that topped clean-context
OOD detection collapse once the context is contaminated. GMM falls from $87.9$ to $65.9$, a $22$-point drop that matches
the intuition that a single global mixture is dragged toward the very outliers
it should flag; TabICL falls from $89.2$ to $73.1$, TACTIC-C from $88.7$ to
$70.7$, and KNN from $87.1$ to $70.0$. The methods that lead here, TACTIC-N, MAF,
RealNVP, and Isolation Forest, were only mid-pack on the clean
task. \our{} is the one method that stays in the top three across both settings.
We attribute this consistency to two design choices rather than to the
in-context formulation alone. The prior, which spans heavy-tailed and perturbed
distributions, exposes the model during pretraining to exactly the low-density
and near-manifold structure that contamination introduces, and the ranking term
in the loss makes the resulting energy ordering robust when absolute values are
imperfect. The result is a single density that scores well whether the context
is clean or contaminated, rather than a method tuned for one regime. It is worth noting that TabEBM, the other energy-based tabular model, scores
near chance on both detection tasks ($52.9$ and $48.7$).

\begin{table*}[t]
\centering
\footnotesize
\caption{\textbf{Prior ablation.} Pairwise ordering accuracy on held-out
synthetic data (higher is better) when each ingredient of the prior is removed
during pretraining. Rows are grouped by the evaluation
distribution class, with \cmark/\xmark\ indicating whether the class includes
categorical features (Cat.) and input perturbations (Aug.). \textbf{Bold} is
best, \underline{underline} second best.}
\label{tab:ablation-prior}
\begin{tabular}{lcc|cccccc}
\toprule
\multicolumn{3}{c|}{Evaluation datasets} & \multicolumn{6}{c}{Prior} \\
Base class & Cat. & Aug. & w/o h-tail & w/o Gauss & w/o flow & w/o cat & w/o augm & full prior \\
\midrule
Gaussian mix. & \cmark & \xmark & \textbf{0.7894} & 0.7772 & 0.7858 & 0.7562 & \underline{0.7869} & 0.7824 \\
Heavy-tailed mix. & \cmark & \xmark & 0.7315 & \underline{0.7545} & 0.7537 & 0.7367 & \textbf{0.7589} & 0.7462 \\
% Random mix. (cat) & 0.7252 & 0.7327 & \textbf{0.7360} & 0.7237 & \underline{0.7345} & 0.7306 \\
% Flow-transformed (cont) & 0.7631 & 0.7708 & 0.7624 & \underline{0.7718} & \textbf{0.7752} & 0.7671 \\
Mix. with flows & \cmark & \xmark & 0.7510 & \underline{0.7543} & 0.7462 & 0.7371 & \textbf{0.7609} & 0.7541 \\
 All datasets & \xmark & \cmark & 0.7354 & 0.7378 & 0.7340 & \textbf{0.7415} & 0.7280 & \underline{0.7381} \\
All datasets & \cmark & \cmark & \underline{0.7188} & 0.7183 & 0.7168 & 0.7010 & 0.7088 & \textbf{0.7209} \\
% \midrule
% Mean & 0.7449 & \underline{0.7494} & 0.7478 & 0.7383 & \textbf{0.7505} & 0.7485 \\
\bottomrule
\end{tabular}
\end{table*}

\begin{figure}[t]
    \centering
    \includegraphics[width=0.9\columnwidth]{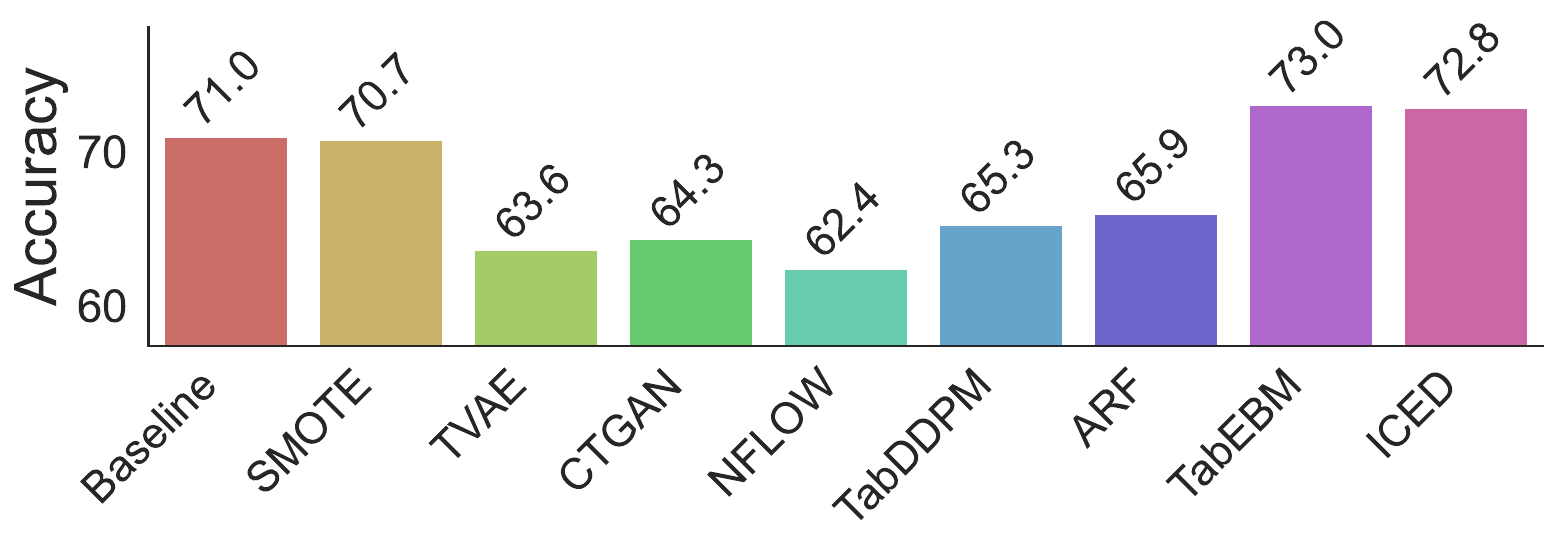}
    \caption{Aggregated downstream
    classification accuracy for classifiers trained on the original data and on data
    augmented by each method (higher is better).
    }
    \label{fig:gen}
\end{figure}
% =====================================================================
\subsection{Data augmentation}
\label{sec:exp-aug}

% \paragraph{Setup.}
The final task is generative rather than discriminative. Following the
TabEBM protocol~\citep[Section~3.1]{margeloiu2024tabebm}, we treat each class as
its own distribution, condition \our{} on the within-class context, and synthesize
new points by following the gradient of its energy with Stochastic Gradient
Langevin Dynamics (SGLD)~\citep{welling2011bayesian}. The augmented set trains multiple
downstream classifiers (see Appendix~\ref{app:augmentation_exp} for details), whose accuracy we compare against training on the
original data alone (\emph{baseline}) across training-set sizes.

% \paragraph{Results.}
In downstream classification accuracy (Figure~\ref{fig:gen}), \our{}
reports $72.8$, essentially matching the purpose-built TabEBM ($73.0$) and
improving over the vanilla baseline ($71.0$), while reusing the very same frozen
model that produced every other result in this paper. Both energy-based methods
clearly outperform the dedicated generative augmenters, several of which fall below the vanilla baseline. 
% The two
% methods use an identical augmentation procedure: both are training-free, both
% condition on a single class at a time so that new points can be generated for a
% chosen class, and both draw those points by running SGLD on an energy landscape.
% The only difference is where the energy comes from. TabEBM obtains it
% indirectly, by converting the logits of a frozen TabPFN classifier on a
% surrogate binary task into a class-specific EBM, whereas \our{} uses the density
% it already models for the conditioning class. That the same frozen density model
% matches a purpose-built augmentation method, with no change of weights and no
% surrogate construction, is the generative face of the one-model-many-tasks
% claim.

% =====================================================================
\subsection{Ablation study}
\label{sec:exp-ablation}

\our{}'s behavior is determined by two design choices, what the prior contains
and what the loss rewards. We ablate each on the synthetic density task of
Section~\ref{sec:exp-density}.

% \paragraph{The prior is the inductive bias.}
First, we retrain \our{} while removing one ingredient of the prior at a time and
evaluate on the held-out synthetic classes
(Table~\ref{tab:ablation-prior}).
Removing a component tends to help slightly on classes that do not exercise it
and to hurt on the combined classes that do. On the hardest setting, evaluation
data with both categorical features and perturbations, the full prior (all) is
best (0.7209), and it is also strongest or second-strongest on the aggregate
rows spanning all datasets. 
% Individual removals occasionally win on a narrow
% class, for example dropping heavy-tailed components slightly raises accuracy on
% the pure Gaussian class, but no ablation matches the full prior once the
% evaluation mixes several structures at once. %This is the signature we expect if
% each component contributes the coverage its name suggests, and if breadth in the
% prior is what lets a single model generalize across distribution families.

% \paragraph{The loss encodes how density is used.}
Next, we compare the full objective against two reduced variants
(Table~\ref{tab:ablation-loss}). The full loss is best on every evaluation class. Dropping the
ranking term costs a consistent margin, confirming that aligning the
model with the order density induces helps even a rank-based metric. Removing
the regression term is more damaging still, leaving the weakest variant
throughout, since the pairwise objective alone does not anchor the energy where
the data concentrate. 
% The two terms are therefore complementary: regression
% supplies the accurate high-density fit and the ranking term sharpens the
% ordering the downstream tasks rely on.

\begin{table}[t]
\centering
\setlength{\tabcolsep}{4pt}
\footnotesize
\caption{\textbf{Loss ablation.} Pairwise ordering accuracy on held-out
synthetic data (higher is better) when each term of the training objective is
removed. Rows are
grouped by evaluation distribution class, with \cmark/\xmark\ indicating whether
the class includes categorical features (Cat.) and input perturbations (Aug.).
\textbf{Bold} is best, \underline{underline} second best.}
\label{tab:ablation-loss}
\begin{tabular}{lcc|ccc}
\toprule
\multicolumn{3}{c|}{Evaluation datasets} & \multicolumn{3}{c}{Loss} \\
Base class & Cat. & Aug. & w/o rank & w/o MSE & full loss \\
\midrule
Gaussian mix.  & \cmark & \xmark & \underline{0.7716} & 0.7541 & \textbf{0.7824} \\
Heavy-tailed mix. & \cmark & \xmark & \underline{0.7389} & 0.7301 & \textbf{0.7462} \\
Mix. with flows & \cmark & \xmark & \underline{0.7449} & 0.7340 & \textbf{0.7541} \\
All datasets & \xmark & \cmark & \underline{0.7299} & 0.7099 & \textbf{0.7381} \\
All datasets & \cmark & \cmark & \underline{0.7107} & 0.6970 & \textbf{0.7209} \\
\bottomrule
\end{tabular}
\end{table}

% =====================================================================
\section{Limitations}
\label{sec:limitations}

First, the accuracy of ICED's density estimation depends on how well the synthetic prior reflects the target
distribution, and we give no formal coverage guarantee. 
However, our experiments show that \our{} transfers well to the OOD, anomaly, and augmentation benchmarks, where the targets are
real datasets outside any prior class. Second, \our{} returns an unnormalized energy, which is sufficient for
ranking, detection, and sampling but cannot serve as an absolute normalized
probability without an extra normalization step. Third, a single forward pass
attends over the whole context, so inference cost grows with context size; this
can be mitigated by subsampling, and recent long-context architectures make
larger contexts increasingly practical. Finally, our augmentation results use
SGLD sampling from the learned energy, which reintroduces an iterative procedure
at generation time even though density estimation remains a single forward pass.

% =====================================================================
\section{Conclusion}
\label{sec:conclusion}

We introduced \our{}, an in-context energy-based density estimator for tabular
data. Rather than fitting a new model to every dataset, \our{} learns the
density-estimation algorithm itself during a single pretraining phase on a
synthetic prior with exact log-densities, and then estimates the energy of any
query point on an unseen dataset in one forward pass, with no retraining,
sampling, or hyperparameter tuning. By modeling an unnormalized energy and
training it with an objective that fits density where it is informative and
preserves its ordering elsewhere, a single frozen model supports the full family
of tasks that density underlies. Across density estimation, out-of-distribution
detection, unsupervised anomaly detection, and data augmentation, \our{} is
competitive with strong task-specific baselines while being the only method that
moves between these tasks without retraining, tuning, or labels. We view \our{}
as a step toward treating density estimation as a reusable, amortized primitive
for tabular data, in the same way that prior-fitted networks have reshaped
supervised tabular learning. 
% Extending the prior to broader classes of
% distributions, scaling in-context estimation to larger contexts, and producing
% calibrated normalized densities are natural directions for future work.

\section*{Acknowledgments}
The research of P. Marszałek and M. Śmieja was supported by the National Science Centre (Poland), grant no. 2023/50/E/ST6/00169. The research of J. Tabor was supported by the National Science Centre (Poland), grant no. 2023/49/B/ST6/01137. Some experiments were performed on servers purchased with funds from the flagship project entitled “Artificial Intelligence Computing Center Core Facility” from the DigiWorld Priority Research Area within the Excellence Initiative – Research University program at Jagiellonian University in Krakow. We gratefully acknowledge Polish high-performance computing infrastructure PLGrid (HPC Center: ACK Cyfronet AGH) for providing computer facilities and support within computational grant no. PLG/2025/018969.

\appendix

\begin{table*}[t]
\centering
\small
\caption{Evaluation of density estimation quality on unseen synthetic data. We measure the agreement between estimated and reference densities using Kendall's $\tau$.}
\label{tab:density-kt}
\begin{tabular}{lcc|ccccccccc}
\toprule
 \multicolumn{3}{c|}{Evaluation datasets} & \multicolumn{9}{c}{Methods} \\
Base class & Cat. & Aug. & KDE & GMM-BIC & MAF & RealNVP & VAE & TabPFN & TabICL & TabEBM & \our{} \\
\midrule
Gaussian mix. & \cmark & \xmark & 0.4710 & \underline{0.5459} & 0.4503 & 0.4677 & 0.4710 & 0.4973 & 0.5342 & 0.1781 & \textbf{0.5648} \\
Heavy-tailed mix. & \cmark & \xmark  & 0.3740 & \underline{0.4482} & 0.3587 & 0.3387 & 0.3000 & 0.4278 & 0.4411 & 0.2025 & \textbf{0.4925} \\
% Mix. w. cat. & 0.6915 & \underline{0.7211} & 0.6824 & 0.6754 & 0.6536 & 0.6844 & 0.6113 & 0.5983 & \textbf{0.7306} \\
% Mix. flow (cont) & 0.7229 & 0.7441 & 0.7163 & 0.7115 & 0.6874 & \underline{0.7464} & 0.7393 & 0.6022 & \textbf{0.7671} \\
Mix. with flows & \cmark & \xmark & 0.3951 & 0.4184 & 0.3971 & 0.3949 & 0.3712 & 0.4379 & \underline{0.4645} & 0.1156 & \textbf{0.5082} \\
All datasets & \xmark & \cmark & 0.3895 & \underline{0.4757} & 0.3602 & 0.3735 & 0.3187 & 0.4218 & 0.4387 & 0.1549 & \textbf{0.4761} \\
All datasets & \cmark & \cmark & 0.3340 & \underline{0.4191} & 0.3243 & 0.2935 & 0.2615 & 0.3772 & 0.4004 & 0.1663 & \textbf{0.4418} \\
% \midrule
% Mean & 0.6984 & \underline{0.7313} & 0.6920 & 0.6887 & 0.6717 & 0.7117 & 0.6812 & 0.5866 & \textbf{0.7485} \\
\bottomrule
\end{tabular}
\end{table*}

\section{Architecture details}
\label{app:architecture}
% TODO: full model architecture (embedding, attention blocks, output head).

\our{} is implemented as a transformer-based in-context learning model that
operates on sets of tabular samples. Each input sample is first mapped into a
token representation using the TabICLv2 feature extractor~\citep{qu2026tabiclv2}. The resulting token embeddings have a dimensionality of $512$.
The feature extractor supports inputs with varying numbers of features and is
initialized from a pretrained TabICLv2 checkpoint. During pretraining, these
weights are fine-tuned jointly with the remaining model parameters.

The resulting tokens are processed by a transformer encoder consisting of
$12$ layers with hidden embedding dimension $d_{\text{model}}=512$. Each
transformer layer uses multi-head self-attention with $4$ attention heads and a
feed-forward network with hidden dimension $1024$, using GELU as the activation
function. Dropout is not applied ($p=0$).
The model processes context and query samples jointly, using an attention mask
that allows query tokens to attend to all context tokens, while restricting
context tokens to attend only to other context tokens. This design preserves
the independence of the context representation from the query set and ensures
permutation invariance over input samples.

The output of the transformer encoder is passed through a lightweight decoder
implemented as a two-layer multilayer perceptron. The decoder first projects the
token representation from the transformer embedding dimension to a hidden
dimension of $1024$ using a linear layer followed by a GELU activation. A final
linear projection maps the hidden representation to the output dimension,
producing a scalar energy estimate for each query sample.

We pretrain each model for $10$M iterations, sampling a new synthetic
dataset from the predefined prior distribution at every iteration. We optimize
the model using the Adam optimizer with a learning rate of
$3\times10^{-5}$. To stabilize training and enable efficient optimization, we
use gradient accumulation over $25$ steps. Before being passed to the model, we normalize all
input features using min-max scaling.

% \section{Prior details}
% \label{app:prior}
% TODO: full sampling distributions and hyperparameters for the synthetic prior.

% Flowy  + augmentacje
% Normalizinf flows: ...

\section{Random pointwise normalizing flows for synthetic data}
\label{app:flows}

To generate distributions with curved, multimodal support while retaining exact
log-densities, we optionally compose the base mixture with a random normalizing
flow. In addition to RealNVP, we use pointwise flows, which is  composed of $L$ layers. A layer $T$ applies an orthogonal
mixing step followed by a pointwise nonlinearity, and for each flow instance the
nonlinearity is drawn uniformly at random from three families, piecewise-linear,
ELU-based, and softplus-based, each with probability $1/3$. Because every
transformation is invertible with a tractable Jacobian, the change-of-variables
rule propagates the exact log-density from the base mixture to the warped
distribution, giving the ground-truth targets used to pretrain \our{}.

% \subsection{Flow layer structure}

Let $z \in \mathbb{R}^d$ be the input to a layer. The layer first applies an
orthogonal mixing map and then a coordinatewise nonlinearity.

\paragraph{Orthogonal mixing.}
For numerical stability and a trivial Jacobian, mixing uses an orthogonal matrix
$Q \in \mathbb{R}^{d \times d}$ with $Q^\top Q = I$,
\begin{equation}
  z' = Q z .
\end{equation}
Since $|\det Q| = 1$, this step contributes nothing to the log-determinant,
$\log|\det J_{\text{lin}}| = 0$, so the entire Jacobian of a layer comes from the
pointwise nonlinearity.

\paragraph{Piecewise-linear activity}

Each coordinate $i$ is transformed around a random breakpoint $k_i$ with distinct
positive slopes on either side,
\begin{equation}
  f_i(z'_i) =
  \begin{cases}
    a_i\,(z'_i - k_i) + k_i, & z'_i \le k_i, \\
    b_i\,(z'_i - k_i) + k_i, & z'_i > k_i,
  \end{cases}
\end{equation}
with $a_i, b_i > 0$. The map is piecewise linear and strictly increasing, and its
log-determinant is
\begin{equation}
  \log|\det J_f| = \sum_{i=1}^{d}
  \log\!\big( a_i\,\mathbb{I}(z'_i \le k_i) + b_i\,\mathbb{I}(z'_i > k_i) \big).
\end{equation}

% \paragraph{Smooth nonlinearities}

The remaining two families replace the sharp knee of the piecewise-linear map
with a smooth transition, which produces densities without hard creases.

\paragraph{ELU-based.}
The exponential linear unit scales negative values nonlinearly while staying
linear for positive ones,
\begin{equation}
  f_i(z'_i) =
  \begin{cases}
    \alpha_i\,(\exp(z'_i - k_i) - 1) + k_i, & z'_i \le k_i, \\
    \beta_i\,(z'_i - k_i) + k_i, & z'_i > k_i,
  \end{cases}
\end{equation}
with $\alpha_i, \beta_i > 0$. On the lower branch the derivative is
$\alpha_i \exp(z'_i - k_i)$ and on the upper branch it is $\beta_i$, so
\begin{multline}
  \log|\det J_f| = \\ \sum_{i=1}^{d}
  \log\!\big( \alpha_i \exp(z'_i - k_i)\,\mathbb{I}(z'_i \le k_i)
  + \beta_i\,\mathbb{I}(z'_i > k_i) \big).
\end{multline}

\paragraph{Softplus-based.}
Softplus is a smooth approximation to ReLU, parameterized here to control the
sharpness of the bend and the residual slope,
\begin{equation}
  f_i(z'_i) = \frac{s_i}{\gamma_i}\,
  \log\!\big(1 + \exp(\gamma_i (z'_i - k_i))\big) + m_i\, z'_i,
\end{equation}
where $\gamma_i$ sets the steepness of the transition and $s_i, m_i$ are random
slope coefficients (with $s_i, m_i > 0$ to keep the map increasing). Its
derivative is a shifted sigmoid, giving
\begin{multline}
  \frac{\partial f_i}{\partial z'_i}
  = s_i\,\sigma\!\big(\gamma_i (z'_i - k_i)\big) + m_i,
  \\
  \log|\det J_f| = \sum_{i=1}^{d}
  \log\!\left( \frac{s_i}{1 + \exp(-\gamma_i (z'_i - k_i))} + m_i \right).
\end{multline}

\paragraph{Change of variables}

Stacking $L$ layers and letting $x \sim p_x$ be a sample from the base Gaussian
mixture, the log-density of the warped output $y$ is obtained by accumulating the
per-layer log-determinants,
\begin{equation}
  \log p_y(y) = \log p_x(x) - \sum_{l=1}^{L} \log|\det J_l| .
\end{equation}
Because each term on the right is available in closed form, every flow-warped
training instance carries an exact ground-truth log-density, which is precisely
what \our{}'s regression target requires.

\section{Query augmentation details}
\label{app:augmentation-details}

During pretraining, we apply query-level augmentations to regularize the
learned energy landscape in regions outside the data manifold. For each
training episode, augmentation is applied with probability $0.5$. If an
augmentation is selected, a random mask is sampled over the query points,
where each coordinate is independently selected with probability $0.1$.
Only the selected coordinates are modified, while the remaining features
remain unchanged.

We consider three types of query perturbations: Gaussian noise injection,
Mixup, and CutMix. When the random augmentation mode is enabled, one of these
transformations is sampled uniformly for each episode.

For Gaussian noise augmentation, the selected query coordinates are perturbed
with zero-mean Gaussian noise. The noise magnitude is adapted to the input
dimensionality and is defined as $\sigma(d) = 7.76 \cdot d^{-1.955}$, where $d$ denotes the number of input features.

For Mixup augmentation, each query sample is interpolated with a randomly
selected context sample. The interpolation coefficient is sampled as
\[
\lambda \sim \mathcal{N}(0.5, 0.1),
\]
and the resulting perturbation is computed as
\[
\Delta x =
\lambda x_{\mathrm{context}} +
(1-\lambda)x_{\mathrm{query}}
-
x_{\mathrm{query}}.
\]
The perturbation is applied only to the coordinates selected by the
augmentation mask.

For CutMix augmentation, a subset of features is replaced with the
corresponding features from randomly sampled context points. The fraction of
replaced features is sampled uniformly from $[0.1,0.5]$. Letting
$\mathcal{I}$ denote the selected feature indices, the perturbed query is
obtained by replacing
$x_{\mathrm{query},\mathcal{I}}$ with
$x_{\mathrm{context},\mathcal{I}}$.

These augmentations generate partially perturbed queries located around the
original data manifold. By exposing the model to such samples during
pretraining, the learned energy function is encouraged to form smoother and
more meaningful landscapes in regions relevant for downstream
out-of-distribution detection.

% =====================================================================

\section{Additional density-estimation results}
\label{app:density-kendall}
% TODO: Kendall's tau table mirroring Table~\ref{tab:density-pa}.

As an additional evaluation metric to the pairwise accuracy reported in Section~4.1, we provide density estimation results using Kendall's $\tau$ correlation coefficient. We evaluate the models under the same synthetic dataset setting as in the main experiments, measuring the rank agreement between the estimated and reference density values. Kendall's $\tau$ is defined as:
\begin{equation}
\tau = \frac{C-D}{\frac{1}{2}n(n-1)},
\end{equation}
where $C$ and $D$ denote the number of concordant and discordant pairs, respectively, among all possible pairs of $n$ samples. Higher values of $\tau$ indicate better agreement between the estimated density ordering and the ground-truth density ordering. The results are presented in Table~\ref{tab:density-kt}.

\FloatBarrier

\section{Extended anomaly detection results}

This section presents the complete results on the ADBench benchmark for both out-of-distribution and unsupervised anomaly detection settings described in Sections~4.2 and~4.3 of the main paper.

In addition to density-based models, we evaluate \our{} against the 14 dominant algorithms from the ADBench~\citep{han2022adbench} benchmark, implemented using the  PyOD~\citep{pyod} library, including KNN~\citep{knn}, LOF~\citep{lof}, IForest~\citep{isoforest}, PCA~\citep{pca}, HBOS~\citep{hbos},  OCSVM~\citep{ocsvm}, CBLOF~\citep{cblof}, DeepSVDD~\citep{deepsvdd}, ECOD~\citep{ecod}, COPOD~\citep{copod}, LODA~\citep{loda}, SOD~\citep{sod}, COF~\citep{cof} and DAGMM~\citep{zong2018deep}. We further extend the comparison with two recent methods, namely TCCM~\citep{tccm} and TACTIC-N (pretrained on clean context) and TACTIC-N (pretrained on noisy context) \citep{tactic2026}. For all baselines, we use the hyperparameter settings recommended by the respective authors to ensure fair comparison.

The reported values are averaged over five random seeds. In addition to the results for individual datasets, Tables~\ref{tab:ood} and~\ref{tab:anomaly} include two aggregation rows: Mean AUROC and Median Rank.

%\begin{figure}[t]
%    \centering
%    \includegraphics[width=\columnwidth]{figs/anomalies/AUCROC_anomalies_False.pdf}
%    \caption{Out-of-distribution detection}
%    \label{fig:an-auroc}
%\end{figure}

%\begin{figure}[t]
%    \centering
%    \includegraphics[width=\columnwidth]{figs/anomalies/AUCROC_anomalies_True.pdf}
%    \caption{Unsupervised anomaly detection...}
%    \label{fig:ood-auroc}
%\end{figure}

\begin{sidewaystable*}[t]
\centering
\tiny
\caption{Out-of-distribution detection AUROC scores on the ADBench benchmark. \textbf{Bold} and \underline{underlined} values denote the best and second-best results, respectively.}
\label{tab:ood}
\resizebox{\textwidth}{!}{
\begin{tabular}{lcccccccccccccccccccccccccc}
\toprule
Idx & TabICL & TACTIC-C & ICED & GMM & KNN & TACTIC-N & MAF & TabPFN & NVP & LOF & KDE & TCCM & IForest & VAE & PCA & HBOS & OCSVM & DeepSVDD & CBLOF & COPOD & ECOD & DAGMM & LODA & SOD & TabEBM & COF \\
\midrule
1 & \textbf{71.76} & 61.58 & 62.03 & 55.86 & 63.05 & 59.08 & 56.10 & 55.98 & 57.10 & \underline{66.84} & 55.82 & 57.11 & 55.76 & 57.82 & 56.34 & 54.49 & 55.17 & 55.46 & 54.63 & 53.36 & 55.21 & 51.87 & 51.06 & 60.28 & 43.56 & 63.28 \\
2 & 92.68 & 88.29 & 93.33 & 94.33 & 87.04 & 91.54 & \underline{94.52} & \textbf{94.64} & 86.68 & 90.92 & 65.46 & 76.64 & 91.44 & 82.43 & 81.23 & 70.78 & 66.03 & 85.28 & 63.83 & 77.56 & 78.97 & 60.84 & 60.61 & 71.39 & 40.56 & 64.30 \\
3 & \textbf{96.87} & 94.92 & 90.66 & \underline{95.99} & 94.91 & 84.20 & 93.76 & 51.55 & 92.52 & 93.99 & 93.87 & 93.24 & 75.74 & 89.35 & 69.33 & 66.49 & 85.79 & 58.25 & 76.73 & 78.24 & 84.50 & 57.32 & 64.50 & 67.60 & 73.75 & 76.62 \\
4 & 99.45 & 99.37 & 99.40 & \textbf{99.69} & 99.49 & 99.55 & 96.40 & 75.75 & 99.12 & 72.64 & 99.47 & \underline{99.56} & 99.54 & 99.39 & 99.16 & 99.49 & 99.43 & 98.20 & 99.22 & 99.50 & 98.97 & 95.54 & 98.48 & 84.55 & 98.47 & 45.03 \\
5 & \underline{81.69} & 78.62 & \textbf{81.78} & 77.47 & 74.40 & 77.45 & 78.46 & 71.53 & 77.03 & 67.18 & 70.44 & 75.83 & 74.05 & 77.80 & 76.87 & 80.33 & 69.42 & 73.26 & 52.54 & 78.31 & 76.91 & 52.18 & 54.67 & 65.89 & 42.96 & 53.41 \\
6 & 93.11 & 92.86 & 90.94 & 84.38 & 95.79 & 92.19 & 87.23 & 89.90 & 95.86 & 95.24 & \underline{97.46} & \textbf{98.11} & 94.85 & 96.40 & 96.70 & 86.82 & 97.45 & 93.77 & 87.88 & 92.28 & 93.41 & 86.03 & 92.22 & 50.32 & 91.64 & 52.91 \\
7 & 78.53 & 71.52 & 72.19 & 66.59 & 79.78 & 76.58 & 65.44 & 80.12 & 78.29 & 80.12 & 82.20 & 81.57 & 80.05 & 81.88 & 82.24 & 70.35 & \underline{84.27} & 79.47 & 79.11 & 66.82 & 78.07 & 74.49 & 78.70 & 36.06 & \textbf{91.84} & 41.14 \\
8 & 60.04 & 74.72 & 76.01 & 78.96 & 65.04 & \textbf{86.23} & \underline{81.45} & 57.90 & 76.86 & 45.99 & 66.60 & 69.26 & 70.48 & 77.45 & 78.87 & 75.94 & 70.16 & 72.55 & 69.21 & 74.55 & 75.11 & 52.60 & 64.03 & 49.33 & 66.01 & 40.94 \\
9 & \textbf{78.04} & 64.24 & 72.68 & 72.18 & 68.23 & 62.99 & \underline{73.26} & 53.71 & 68.13 & 50.27 & 68.10 & 61.50 & 62.96 & 70.86 & 71.70 & 65.65 & 56.28 & 70.32 & 54.44 & 67.94 & 66.54 & 56.62 & 41.95 & 58.76 & 49.03 & 40.13 \\
10 & 95.31 & 94.18 & 95.14 & 92.32 & 97.60 & 82.88 & 87.29 & \underline{98.61} & 96.20 & \textbf{99.06} & 91.00 & 91.91 & 84.65 & 95.96 & 94.44 & 76.68 & 95.21 & 87.64 & 89.93 & 87.90 & 91.56 & 87.46 & 92.47 & 59.56 & 9.44 & 51.55 \\
11 & 98.79 & \underline{99.43} & 96.58 & 98.18 & \textbf{99.83} & 96.91 & 97.43 & 86.12 & 97.28 & 95.48 & 95.07 & 93.63 & 90.11 & 88.59 & 88.52 & 82.36 & 88.18 & 83.72 & 83.41 & 81.49 & 88.95 & 69.71 & 44.62 & 58.63 & 85.24 & 70.83 \\
12 & 73.91 & 78.40 & \textbf{81.09} & 77.74 & \textbf{81.09} & 70.45 & 71.00 & 79.35 & 74.55 & 66.04 & 78.05 & 72.17 & 66.53 & 58.94 & 55.07 & 63.17 & 55.63 & 55.52 & 62.76 & 44.78 & 46.64 & 46.52 & 50.74 & 52.78 & 55.01 & 54.14 \\
13 & 94.50 & 94.24 & 93.45 & 93.50 & 94.93 & \textbf{95.34} & 92.16 & 92.51 & 93.78 & \underline{95.17} & 93.51 & 93.49 & 93.21 & 93.26 & 93.26 & 94.09 & 93.24 & 92.95 & 88.25 & 92.18 & 92.66 & 86.06 & 92.59 & 94.43 & 11.45 & 86.29 \\
14 & 98.00 & \underline{99.19} & 98.80 & 99.16 & 94.12 & 90.93 & 80.51 & \textbf{99.44} & 75.98 & 91.92 & 77.90 & 87.95 & 81.78 & 73.41 & 74.07 & 81.76 & 49.84 & 67.11 & 72.06 & 75.25 & 71.77 & 79.49 & 74.64 & 60.24 & 70.27 & 48.03 \\
15 & 98.25 & 99.95 & \textbf{100.00} & \textbf{100.00} & 99.78 & 89.38 & 90.44 & \textbf{100.00} & 99.64 & 66.35 & 99.85 & 82.51 & 84.16 & 83.19 & 85.19 & 86.25 & 87.34 & 78.46 & 69.75 & 81.68 & 74.48 & 59.36 & 62.43 & 3.57 & 18.35 & 18.34 \\
16 & 97.53 & 99.91 & 99.55 & \textbf{100.00} & 99.98 & 99.90 & 99.99 & 93.43 & 99.99 & 99.96 & 99.57 & 99.96 & 99.52 & \textbf{100.00} & 99.95 & 99.57 & 99.57 & 99.84 & 84.40 & 99.08 & 97.96 & 99.73 & 37.14 & 57.56 & 14.68 & 4.08 \\
17 & \textbf{90.79} & 82.24 & 61.52 & \underline{90.60} & 82.01 & 72.01 & 88.90 & 50.00 & 75.34 & 85.69 & 87.04 & 70.24 & 41.28 & 88.41 & 79.60 & 53.58 & 70.79 & 77.91 & 54.27 & 67.57 & 67.63 & 63.11 & 56.02 & 35.11 & nan & 45.06 \\
18 & 98.17 & \underline{99.70} & 99.41 & \textbf{99.82} & 97.54 & 96.70 & 97.09 & 99.43 & 97.96 & 92.63 & 96.98 & 98.07 & 90.97 & 90.75 & 89.56 & 77.46 & 92.41 & 85.07 & 88.92 & 79.81 & 72.66 & 75.32 & 87.12 & 51.58 & 98.68 & 66.06 \\
19 & 63.86 & \textbf{86.25} & 72.93 & 72.65 & 77.72 & \underline{82.84} & 63.74 & 78.45 & 52.52 & 76.47 & 58.27 & 58.67 & 62.91 & 50.58 & 41.55 & 69.71 & 38.87 & 43.70 & 65.77 & 43.21 & 37.54 & 49.62 & 42.34 & 45.42 & 28.12 & 51.15 \\
20 & \underline{92.49} & 92.09 & 86.58 & \textbf{95.43} & 91.12 & 84.60 & 86.01 & 91.12 & 80.99 & 89.35 & 73.72 & 87.14 & 63.82 & 54.16 & 53.73 & 59.09 & 51.84 & 50.38 & 67.95 & 55.74 & 57.49 & 53.80 & 54.14 & 76.44 & 37.44 & 80.84 \\
21 & 98.13 & \textbf{100.00} & \textbf{100.00} & \textbf{100.00} & \textbf{100.00} & 99.88 & 99.64 & 99.19 & 99.99 & 97.77 & \textbf{100.00} & \textbf{100.00} & 99.69 & 99.93 & 99.94 & 99.77 & \textbf{100.00} & 99.01 & 80.80 & 99.46 & 99.50 & 91.02 & 41.54 & 37.90 & 79.84 & 31.00 \\
22 & 86.71 & \textbf{87.83} & 86.11 & 86.78 & 85.56 & 78.74 & 86.40 & \underline{87.46} & 82.63 & 80.95 & 76.98 & 77.38 & 77.71 & 71.58 & 70.14 & 73.53 & 71.70 & 67.86 & 74.33 & 68.43 & 64.06 & 73.89 & 72.73 & 66.32 & 23.91 & 59.88 \\
23 & 84.01 & \textbf{91.69} & 86.23 & 87.00 & 88.72 & 85.13 & 85.36 & 85.38 & 88.00 & 85.65 & 89.48 & 88.68 & 88.15 & 87.71 & 90.23 & 85.93 & 90.09 & 85.44 & 74.75 & 90.68 & \underline{90.83} & 84.75 & 88.24 & 78.50 & 29.89 & 68.07 \\
24 & - & \underline{95.45} & 91.03 & 78.97 & 93.55 & 87.45 & 92.20 & 64.06 & 93.07 & \textbf{95.49} & 93.76 & 85.71 & 86.29 & 90.62 & 90.50 & 67.15 & 88.89 & 81.18 & 60.23 & 77.42 & 74.50 & 68.25 & 83.79 & 62.36 & 29.44 & 57.77 \\
25 & 98.72 & \textbf{100.00} & \textbf{100.00} & 99.98 & \textbf{100.00} & \textbf{100.00} & \textbf{100.00} & 99.16 & \textbf{100.00} & \textbf{100.00} & \textbf{100.00} & \textbf{100.00} & 95.61 & \textbf{100.00} & \textbf{100.00} & \textbf{100.00} & \textbf{100.00} & 98.61 & 65.20 & 94.24 & 95.40 & 95.96 & 99.87 & 54.46 & 97.92 & 51.91 \\
26 & - & 96.55 & 87.19 & 96.91 & 99.70 & 84.51 & 80.17 & 95.43 & 89.71 & \textbf{99.75} & \underline{99.71} & 88.59 & 77.48 & 60.80 & 58.86 & 87.78 & 65.30 & 54.81 & 59.48 & 69.40 & 61.49 & 70.98 & 65.95 & 44.44 & 13.50 & 57.40 \\
27 & 92.22 & 93.39 & \textbf{96.89} & 94.21 & 95.25 & 93.46 & 93.94 & 94.11 & 95.97 & \underline{96.09} & 90.83 & 93.37 & 92.40 & 92.63 & 93.14 & 80.85 & 91.93 & 92.63 & 85.12 & 87.66 & 91.37 & 87.93 & 88.38 & 56.00 & 15.85 & 62.04 \\
28 & 98.85 & \textbf{99.92} & 99.29 & 99.39 & \underline{99.91} & 98.32 & 98.69 & 99.50 & 96.26 & 99.23 & 99.25 & 98.68 & 97.45 & 90.24 & 94.39 & 93.93 & 96.56 & 87.21 & 85.97 & 90.51 & 92.85 & 66.54 & 93.32 & 52.22 & 64.89 & 51.39 \\
29 & \underline{87.72} & 84.60 & 85.92 & 68.88 & 79.63 & 79.13 & 78.72 & \textbf{88.66} & 78.16 & 75.05 & 78.15 & 75.27 & 77.55 & 74.77 & 76.19 & 74.79 & 74.83 & 74.63 & 75.78 & 69.72 & 63.49 & 67.28 & 71.13 & 36.69 & 73.75 & 38.83 \\
30 & 80.67 & 87.44 & \underline{87.96} & 84.10 & 87.56 & \textbf{88.94} & 81.32 & 86.98 & 79.39 & 84.85 & 81.52 & 81.72 & 79.27 & 72.71 & 66.30 & 86.45 & 65.54 & 67.64 & 84.64 & 63.27 & 58.31 & 66.18 & 68.59 & 46.94 & 12.22 & 50.95 \\
31 & 98.65 & 97.08 & 99.85 & 99.31 & \underline{99.90} & 99.45 & 99.67 & 99.78 & \textbf{99.92} & 99.70 & 99.73 & 99.88 & 99.28 & 99.53 & 97.84 & 98.13 & 98.16 & 97.39 & 90.09 & 97.42 & 96.46 & 72.57 & 98.35 & 74.58 & 1.54 & 56.38 \\
32 & 98.01 & 99.90 & \underline{99.95} & \textbf{99.97} & 99.86 & 90.20 & \underline{99.95} & 99.19 & 99.74 & 99.77 & 98.63 & 99.87 & 99.64 & \underline{99.95} & 99.58 & 99.00 & 98.55 & 99.50 & 97.88 & 99.47 & 99.22 & 98.54 & 90.78 & 41.27 & 4.32 & 56.89 \\
33 & 97.56 & 91.62 & 96.55 & 94.73 & \underline{98.72} & 92.74 & 93.86 & \textbf{98.88} & 92.56 & 77.55 & 74.47 & 61.42 & 89.45 & 74.31 & 53.52 & 78.35 & 76.94 & 62.51 & 82.72 & 47.60 & 48.91 & 89.93 & 74.94 & 42.33 & 29.18 & 41.64 \\
34 & 91.53 & 90.49 & 92.66 & 93.85 & \underline{95.42} & \textbf{95.95} & 93.15 & 80.22 & 91.50 & 90.95 & 91.93 & 92.69 & 93.67 & 91.06 & 93.88 & 92.09 & 91.80 & 90.58 & 86.64 & 91.21 & 93.63 & 87.11 & 71.91 & 88.69 & 78.58 & 88.60 \\
35 & 79.53 & 72.17 & 75.25 & 76.68 & 70.16 & 66.99 & 77.45 & 64.93 & 69.47 & 73.99 & 65.28 & \textbf{81.04} & 75.21 & \underline{80.08} & 71.52 & 65.80 & 65.73 & 70.85 & 74.90 & 61.33 & 57.36 & 59.41 & 66.48 & 21.25 & 63.25 & 16.87 \\
36 & \underline{62.83} & 48.51 & 52.05 & 50.94 & 48.99 & 48.62 & 47.87 & 48.59 & 49.00 & 50.47 & \textbf{70.79} & 51.14 & 47.33 & 47.34 & 47.10 & 47.38 & 46.57 & 51.55 & 54.09 & 49.19 & 47.05 & 51.04 & 47.08 & 43.85 & 47.08 & 37.63 \\
37 & 98.48 & 98.62 & 98.55 & \underline{99.00} & 96.02 & 95.65 & 93.12 & \textbf{99.42} & 92.69 & 92.94 & 91.78 & 89.83 & 93.17 & 90.51 & 92.69 & 92.44 & 90.84 & 93.93 & 85.60 & 92.89 & 86.87 & 88.33 & 90.47 & 37.38 & 97.25 & 41.01 \\
38 & 98.35 & 94.90 & \textbf{99.08} & 98.12 & 98.14 & 96.06 & 98.60 & 98.11 & 98.33 & 98.28 & 93.87 & 97.54 & \underline{98.94} & 98.27 & 98.12 & 98.68 & 94.11 & 98.50 & 86.52 & 93.79 & 97.74 & 84.45 & 91.86 & 88.14 & 51.89 & 52.63 \\
39 & \textbf{98.59} & 83.82 & 77.30 & 64.26 & 64.38 & 71.06 & 49.45 & \underline{98.26} & 46.83 & 63.37 & 38.91 & 54.58 & 46.64 & 43.28 & 46.69 & 38.31 & 40.35 & 42.72 & 47.54 & 24.27 & 41.62 & 48.27 & 34.07 & 17.53 & 62.66 & 21.32 \\
40 & 98.73 & \underline{99.86} & 97.51 & 99.69 & 98.28 & 95.35 & 92.47 & \textbf{99.91} & 93.09 & 97.26 & 74.30 & 94.29 & 77.57 & 63.92 & 58.16 & 71.02 & 62.78 & 58.53 & 77.70 & 50.30 & 59.54 & 62.77 & 72.76 & 75.26 & 30.10 & 78.90 \\
41 & 77.21 & 73.41 & \underline{86.58} & 78.88 & 77.77 & \textbf{87.05} & 63.41 & 75.92 & 62.08 & 77.27 & 77.30 & 62.81 & 72.72 & 70.89 & 65.91 & 70.42 & 56.82 & 58.94 & 72.06 & 73.82 & 60.97 & 56.15 & 62.39 & 56.32 & 47.32 & 61.40 \\
42 & 98.95 & 99.90 & \textbf{100.00} & \textbf{100.00} & 99.22 & 99.43 & 97.46 & 74.32 & 98.72 & 84.10 & 99.54 & 99.55 & 99.38 & 98.96 & 99.16 & 99.06 & 99.52 & 98.21 & 99.64 & 99.21 & 99.19 & 86.67 & 97.66 & 84.01 & 99.24 & 34.82 \\
43 & 99.44 & \underline{99.99} & 99.77 & 99.75 & 99.77 & 96.09 & 99.53 & 99.77 & 99.34 & 99.90 & 99.87 & 99.65 & 98.98 & 99.70 & 99.46 & 99.10 & 99.79 & 99.09 & 86.72 & 99.39 & 96.73 & 91.23 & 98.95 & 78.35 & \textbf{100.00} & 34.34 \\
44 & 87.23 & 83.16 & 89.18 & 89.36 & 51.87 & 61.32 & \textbf{93.54} & \underline{91.98} & 46.77 & 53.36 & 35.02 & 31.74 & 46.91 & 31.89 & 39.35 & 34.48 & 34.98 & 34.81 & 46.24 & 33.96 & 38.32 & 42.71 & 31.24 & 47.26 & 42.70 & 43.96 \\
45 & 98.80 & \textbf{100.00} & 99.94 & 99.54 & 99.60 & 98.52 & 98.65 & 99.91 & 99.57 & 98.32 & 98.17 & 99.09 & 91.60 & 95.88 & 93.73 & 95.82 & 95.39 & 82.03 & 96.81 & 85.79 & 73.03 & 74.83 & 92.93 & 2.92 & \textbf{100.00} & 45.34 \\
46 & 98.23 & 98.66 & 96.79 & \underline{99.59} & 75.40 & 67.01 & 58.21 & \textbf{99.69} & 59.58 & 61.18 & 57.99 & 53.84 & 57.39 & 55.50 & 53.41 & 60.83 & 53.72 & 52.35 & 54.74 & 53.67 & 50.98 & 55.05 & 53.27 & 7.94 & 44.34 & 20.92 \\
47 & 50.63 & 48.53 & 45.67 & 45.48 & 45.77 & 47.98 & 44.50 & 48.73 & 45.64 & 48.55 & 39.19 & \underline{55.67} & 41.74 & 44.47 & 44.47 & 41.41 & 42.36 & 43.63 & 51.82 & 37.64 & 45.58 & 41.10 & 49.86 & 39.92 & \textbf{64.65} & 34.81 \\
\midrule
Mean & \textbf{89.19} & \underline{88.71} & 88.33 & 87.88 & 87.08 & 85.04 & 84.65 & 84.62 & 83.49 & 83.04 & 82.27 & 82.18 & 80.11 & 79.51 & 77.91 & 77.59 & 76.60 & 75.60 & 74.53 & 74.45 & 74.30 & 70.65 & 70.19 & 53.67 & 52.87 & 50.02 \\
Median-Rank & 6.00 & \textbf{5.00} & \textbf{5.00} & \textbf{5.00} & 6.00 & 10.00 & 10.00 & 7.00 & 10.00 & 8.00 & 11.00 & 10.00 & 12.00 & 13.00 & 14.00 & 15.00 & 16.00 & 18.00 & 16.00 & 18.00 & 18.00 & 21.00 & 19.00 & 23.00 & 24.00 & 24.00 \\
\bottomrule
\end{tabular}
}
\end{sidewaystable*}

\begin{sidewaystable*}[ht]
\centering
\tiny
\caption{Comparison of baselines on the unsupervised anomaly detection task using the ADBench benchmark.}
\label{tab:anomaly}
\resizebox{\textwidth}{!}{
\begin{tabular}{lcccccccccccccccccccccccccc}
\toprule
Idx & TabICL & TACTIC-C & ICED & GMM & KNN & TACTIC-N & MAF & TabPFN & NVP & LOF & KDE & TCCM & IForest & VAE & PCA & HBOS & OCSVM & DeepSVDD & CBLOF & COPOD & ECOD & DAGMM & LODA & SOD & TabEBM & COF \\
\midrule
1 & \textbf{69.27} & 61.46 & 61.29 & 56.17 & 62.55 & 59.86 & 55.99 & 53.69 & 57.01 & \underline{66.95} & 55.49 & 57.37 & 55.93 & 56.84 & 56.05 & 54.04 & 54.90 & 55.21 & 51.96 & 53.20 & 55.48 & 51.28 & 50.64 & 62.07 & 48.33 & 64.11 \\
2 & 86.11 & 70.42 & 88.16 & 88.09 & 71.72 & 76.20 & \underline{88.95} & 88.49 & 71.94 & 70.19 & 57.43 & \textbf{92.68} & 81.99 & 61.39 & 64.94 & 59.88 & 58.04 & 68.27 & 52.55 & 76.95 & 78.53 & 55.42 & 46.86 & 77.92 & 33.64 & 67.30 \\
3 & 87.83 & 88.97 & 71.81 & \textbf{96.37} & 77.53 & 81.02 & \underline{92.85} & 50.35 & 90.75 & 82.96 & 91.98 & 91.17 & 72.75 & 85.62 & 80.90 & 71.57 & 84.63 & 76.92 & 54.95 & 79.52 & 84.95 & 56.42 & 72.44 & 66.89 & 42.75 & 73.06 \\
4 & 93.62 & 64.75 & 96.92 & 93.57 & 97.37 & 95.00 & 94.97 & 95.48 & 97.46 & 41.00 & \underline{99.55} & \textbf{99.56} & 98.33 & 99.31 & 96.43 & 98.67 & 91.81 & 90.33 & 77.34 & \underline{99.55} & 99.21 & 93.42 & 98.53 & 93.55 & 52.76 & 38.67 \\
5 & \underline{79.31} & 74.64 & \textbf{79.64} & 75.24 & 72.02 & 76.14 & 73.74 & 60.91 & 74.03 & 58.96 & 67.00 & 72.15 & 70.85 & 72.37 & 72.56 & 78.32 & 65.87 & 69.11 & 48.49 & 77.77 & 76.08 & 50.37 & 51.09 & 69.58 & 54.59 & 57.61 \\
6 & 64.11 & 68.38 & 75.56 & 65.58 & 75.61 & 87.94 & 79.47 & 57.39 & 90.13 & 66.45 & \underline{95.08} & 67.28 & 92.71 & 93.22 & \textbf{95.27} & 86.53 & 93.32 & 88.74 & 77.50 & 92.43 & 93.47 & 79.19 & 88.25 & 70.31 & 84.84 & 64.67 \\
7 & 52.21 & 49.97 & 50.13 & 47.02 & 55.95 & 63.87 & 48.63 & 41.43 & 63.69 & 59.75 & 75.67 & 78.37 & 68.71 & 70.89 & 75.73 & 60.17 & 78.52 & 73.60 & 61.14 & 66.99 & \underline{78.81} & 64.74 & 73.48 & 51.57 & \textbf{90.22} & 53.25 \\
8 & 58.28 & 71.89 & 71.87 & 78.49 & 60.33 & \textbf{85.36} & \underline{80.19} & 59.40 & 73.95 & 40.76 & 60.65 & 67.03 & 69.88 & 74.43 & 78.22 & 75.03 & 68.67 & 71.00 & 65.04 & 74.75 & 75.52 & 51.66 & 61.92 & 48.54 & 38.64 & 42.00 \\
9 & \textbf{78.95} & 63.93 & 69.70 & 70.14 & 66.14 & 64.56 & \underline{70.59} & 52.73 & 65.60 & 48.04 & 65.64 & 60.37 & 60.29 & 68.86 & 68.20 & 64.04 & 54.56 & 67.45 & 49.88 & 68.46 & 67.03 & 53.52 & 42.45 & 61.64 & 67.80 & 41.55 \\
10 & 82.19 & 56.35 & 88.05 & 89.11 & 87.07 & \textbf{98.09} & 75.41 & 85.88 & 92.42 & 85.94 & 86.31 & 94.56 & 87.10 & \underline{96.08} & 93.63 & 77.15 & 91.84 & 86.32 & 65.13 & 88.17 & 91.51 & 83.03 & 88.08 & 75.09 & 12.04 & 79.80 \\
11 & 85.71 & 70.04 & \textbf{89.74} & 87.75 & 81.86 & 74.01 & 84.03 & 73.91 & 78.08 & 57.07 & 75.18 & 47.23 & 78.40 & 73.41 & 81.86 & 76.40 & 72.37 & 79.11 & 66.40 & 81.33 & \underline{89.01} & 59.74 & 45.57 & 62.93 & 71.57 & 71.01 \\
12 & 58.95 & 64.90 & 65.78 & 67.00 & \textbf{73.84} & 58.48 & 53.75 & 65.91 & 67.09 & 60.20 & \underline{70.28} & 63.41 & 57.70 & 50.93 & 47.16 & 54.24 & 50.28 & 48.45 & 53.64 & 45.04 & 46.73 & 46.24 & 45.25 & 68.39 & 46.87 & 62.66 \\
13 & \underline{99.48} & 95.61 & 92.17 & 91.72 & 95.92 & 96.15 & 93.17 & 91.15 & 95.45 & 97.23 & 94.70 & 93.15 & 94.04 & \textbf{99.52} & 94.13 & 94.01 & 94.55 & 94.00 & 77.20 & 92.87 & 93.29 & 81.42 & 93.96 & 96.40 & 2.02 & 97.63 \\
14 & 64.30 & 73.99 & 82.33 & \underline{82.87} & 82.34 & \textbf{83.52} & 76.25 & 65.06 & 76.56 & 67.84 & 75.49 & 68.64 & 79.44 & 72.84 & 68.88 & 78.68 & 40.20 & 63.33 & 64.40 & 72.93 & 68.41 & 73.32 & 74.86 & 79.21 & 79.11 & 75.93 \\
15 & 37.69 & 72.22 & 77.01 & 40.59 & 51.09 & 78.40 & 74.57 & 38.27 & 70.26 & 39.62 & 70.47 & 75.23 & 70.20 & 72.54 & 75.83 & \underline{78.95} & 66.17 & 70.42 & 55.39 & \textbf{81.78} & 75.43 & 49.16 & 63.12 & 66.88 & 23.64 & 37.86 \\
16 & 99.11 & 99.17 & 99.23 & 7.25 & 5.16 & 99.54 & 99.36 & 41.58 & 82.06 & 29.28 & 99.50 & 72.61 & \textbf{99.97} & 99.26 & 99.60 & 99.45 & 99.50 & \underline{99.66} & 99.20 & 99.07 & 97.97 & 99.53 & 10.07 & 77.82 & 1.81 & 87.66 \\
17 & \underline{73.22} & 65.76 & 56.16 & \textbf{74.58} & 68.97 & 68.36 & 68.43 & 50.00 & 68.97 & 65.30 & 65.88 & 67.76 & 66.79 & 57.00 & 60.59 & 67.49 & 67.72 & 59.56 & 53.17 & 66.39 & 66.45 & 63.62 & 54.36 & 60.98 & nan & 64.41 \\
18 & 71.52 & 73.57 & 87.21 & 78.35 & 89.26 & 90.66 & \textbf{94.18} & 52.13 & 92.87 & 89.88 & 91.73 & \underline{93.14} & 84.44 & 80.94 & 79.26 & 64.63 & 77.84 & 72.66 & 64.09 & 78.90 & 73.42 & 69.06 & 79.05 & 85.84 & 92.23 & 85.83 \\
19 & 51.74 & \underline{67.41} & 55.14 & 59.80 & 58.01 & \textbf{68.14} & 56.04 & 60.07 & 44.94 & 54.29 & 46.27 & 53.75 & 48.47 & 42.43 & 36.33 & 55.78 & 35.97 & 38.88 & 55.36 & 42.01 & 36.57 & 51.16 & 37.48 & 60.03 & 41.28 & 53.30 \\
20 & \textbf{91.64} & 86.65 & 85.92 & \underline{90.46} & 87.16 & 80.63 & 81.92 & 85.42 & 76.48 & 85.48 & 69.50 & 81.36 & 62.95 & 52.41 & 53.00 & 60.94 & 49.91 & 49.71 & 52.81 & 55.78 & 57.83 & 52.22 & 52.30 & 85.49 & 40.76 & 81.35 \\
21 & 76.06 & 98.13 & 99.42 & 23.01 & 64.16 & \textbf{99.89} & 99.02 & 42.08 & 97.15 & 88.64 & 93.43 & 99.80 & \textbf{99.89} & 99.84 & 99.88 & 99.61 & 99.80 & 98.35 & 67.72 & 99.41 & 99.45 & 76.23 & 80.68 & 77.88 & 89.62 & 90.59 \\
22 & 81.15 & 70.81 & 81.87 & 81.26 & \textbf{82.65} & 78.59 & 80.19 & \underline{82.41} & 74.18 & 68.67 & 72.31 & 71.34 & 73.02 & 67.66 & 67.00 & 70.88 & 61.76 & 64.32 & 63.09 & 68.77 & 64.55 & 67.98 & 68.96 & 75.76 & 30.80 & 66.66 \\
23 & 82.09 & 88.77 & 77.47 & 84.79 & 84.41 & 82.76 & 83.51 & 82.36 & 84.45 & 75.83 & 86.47 & 84.08 & 86.24 & 87.26 & 89.69 & 85.63 & 84.78 & 81.59 & 56.19 & \underline{90.74} & \textbf{90.78} & 79.76 & 82.42 & 80.74 & 55.47 & 76.82 \\
24 & - & 85.79 & 78.09 & 64.78 & 80.80 & \underline{86.33} & 85.91 & 58.89 & 85.62 & 67.76 & 78.61 & 77.28 & 81.45 & \textbf{86.49} & 85.75 & 61.23 & 83.01 & 70.43 & 50.10 & 78.18 & 75.23 & 64.24 & 73.08 & 60.81 & 36.25 & 70.98 \\
25 & 78.16 & 91.70 & 90.24 & 31.86 & 68.20 & \textbf{100.00} & 99.91 & 51.38 & 99.43 & 35.49 & 26.65 & 74.51 & 99.99 & \textbf{100.00} & \textbf{100.00} & \textbf{100.00} & 91.84 & 95.66 & 80.71 & 93.89 & 95.03 & 43.31 & 95.09 & 73.34 & 79.22 & 40.85 \\
26 & - & 64.66 & 42.89 & 57.61 & 42.18 & 57.91 & 53.88 & \underline{70.60} & 57.57 & 57.52 & 39.92 & 51.83 & 68.78 & 51.25 & 52.18 & \textbf{83.79} & 55.37 & 50.02 & 50.31 & 68.98 & 60.79 & 68.71 & 59.25 & 57.64 & 10.43 & 50.05 \\
27 & 87.49 & 88.88 & 91.46 & 77.12 & 80.97 & \textbf{95.35} & 88.34 & 82.20 & \underline{92.45} & 74.48 & 89.03 & 81.21 & 89.76 & 75.37 & 90.48 & 79.69 & 89.19 & 89.47 & 68.69 & 87.94 & 91.55 & 86.49 & 82.31 & 75.92 & 31.94 & 72.66 \\
28 & 82.49 & 79.23 & 85.10 & 78.35 & 72.18 & 90.29 & 83.07 & 80.46 & 81.26 & 46.80 & \textbf{96.17} & 79.77 & \underline{95.22} & 83.66 & 93.50 & 92.31 & 94.93 & 85.93 & 73.42 & 90.32 & 92.59 & 64.27 & 91.20 & 66.15 & 86.43 & 43.86 \\
29 & 53.57 & 63.49 & 69.72 & 60.14 & 71.55 & 70.58 & \textbf{74.01} & 59.09 & 71.72 & 63.62 & \underline{72.74} & 71.45 & 71.43 & 66.77 & 69.79 & 70.94 & 66.09 & 67.51 & 55.42 & 69.49 & 63.57 & 65.72 & 66.14 & 60.11 & 66.83 & 58.68 \\
30 & 73.87 & 61.47 & 72.32 & 67.24 & 65.37 & \textbf{82.21} & 72.93 & 64.83 & 64.52 & 56.11 & 68.98 & 64.06 & 70.21 & 67.23 & 59.65 & \underline{74.96} & 59.41 & 61.42 & 63.69 & 63.18 & 58.10 & 59.59 & 61.05 & 64.40 & 24.40 & 55.43 \\
31 & 96.85 & 63.81 & 96.38 & 98.36 & 92.54 & \textbf{99.80} & 98.78 & 86.20 & 99.45 & 49.42 & 99.55 & 85.32 & 99.04 & \underline{99.76} & 96.88 & 97.01 & 97.29 & 96.52 & 80.27 & 96.43 & 95.26 & 77.32 & 97.29 & 84.91 & 1.36 & 55.40 \\
32 & 97.36 & 84.98 & 89.74 & 86.32 & 67.76 & 85.09 & 99.36 & 96.97 & 97.60 & 56.34 & 98.22 & 69.90 & \textbf{99.63} & 99.18 & 98.74 & 98.60 & 97.75 & 98.55 & 83.15 & \underline{99.41} & 99.20 & 97.32 & 62.02 & 69.58 & 0.16 & 53.58 \\
33 & 62.32 & 49.96 & 59.76 & \underline{73.65} & 71.85 & 64.37 & \textbf{76.49} & 72.09 & 73.38 & 46.31 & 61.18 & 61.54 & 67.55 & 50.23 & 42.19 & 60.03 & 50.19 & 48.65 & 49.06 & 47.34 & 48.97 & 73.43 & 45.37 & 59.72 & 38.43 & 41.57 \\
34 & \textbf{100.00} & 86.75 & 90.88 & 91.24 & 93.77 & 95.02 & 88.35 & 70.16 & 91.17 & 83.10 & 88.62 & 93.60 & 92.98 & \textbf{100.00} & 92.09 & 81.98 & 88.37 & 85.92 & 66.94 & 86.71 & 91.35 & 82.64 & 75.73 & 77.39 & \textbf{100.00} & 87.75 \\
35 & 46.91 & 52.56 & 53.34 & 39.93 & 53.68 & 50.61 & 55.08 & 47.92 & 55.19 & 43.06 & 55.93 & 49.63 & 65.38 & 57.09 & 54.92 & 65.14 & 53.43 & 54.52 & 50.42 & \textbf{69.62} & \underline{66.35} & 52.32 & 42.50 & 52.74 & 51.55 & 40.95 \\
36 & \textbf{65.56} & 50.99 & 53.00 & 53.36 & 51.51 & 51.28 & 51.01 & 48.58 & 53.08 & 52.94 & \underline{61.09} & 53.23 & 51.74 & 54.56 & 50.69 & 50.71 & 50.19 & 49.23 & 49.20 & 52.39 & 50.71 & 50.95 & 49.66 & 54.23 & 55.47 & 51.29 \\
37 & 55.95 & 67.86 & 79.17 & 67.14 & 68.44 & 80.61 & 85.62 & 79.35 & 85.15 & 53.22 & 89.73 & 70.58 & 90.15 & 88.00 & 90.21 & 90.38 & 82.98 & \underline{91.32} & 52.46 & \textbf{92.55} & 86.82 & 84.06 & 86.43 & 75.50 & 47.64 & 58.65 \\
38 & \textbf{98.50} & 85.43 & 98.22 & 95.63 & 95.94 & 95.89 & 97.21 & 94.78 & 95.77 & 87.95 & 88.14 & 93.69 & \underline{98.40} & 97.02 & 96.64 & 96.30 & 88.34 & 97.15 & 61.64 & 94.46 & 98.11 & 80.66 & 79.09 & 91.61 & 14.91 & 91.57 \\
39 & \textbf{53.72} & 35.85 & 34.58 & 36.41 & 38.64 & 43.40 & 33.83 & 41.09 & 39.09 & 50.07 & 35.59 & \underline{51.02} & 37.60 & 42.44 & 43.76 & 30.69 & 38.19 & 39.70 & 47.28 & 24.35 & 42.16 & 44.55 & 32.13 & 43.39 & 40.75 & 49.30 \\
40 & 84.90 & 86.30 & 92.72 & 93.04 & \underline{97.28} & \textbf{99.47} & 86.74 & 90.43 & 89.21 & 93.45 & 69.18 & 91.71 & 73.36 & 52.34 & 58.93 & 69.10 & 58.49 & 59.06 & 67.96 & 49.87 & 59.42 & 55.78 & 69.07 & 88.11 & 36.67 & 93.77 \\
41 & 65.16 & 69.05 & \underline{80.43} & 70.24 & 74.27 & \textbf{83.49} & 61.26 & 67.42 & 60.27 & 72.83 & 76.32 & 65.51 & 70.85 & 68.03 & 65.74 & 69.75 & 57.53 & 58.75 & 53.05 & 74.87 & 62.56 & 53.22 & 63.52 & 68.60 & 49.06 & 73.77 \\
42 & 72.08 & 91.45 & 97.09 & 26.51 & 84.93 & 98.58 & 96.21 & 21.36 & 95.25 & 53.87 & \textbf{99.38} & \textbf{99.38} & 99.08 & 98.68 & 98.67 & 98.56 & 99.30 & 98.17 & 90.82 & 99.15 & 99.13 & 91.15 & 97.87 & 94.65 & 93.86 & 52.25 \\
43 & 81.03 & 85.60 & 91.28 & 9.00 & 93.26 & \textbf{99.77} & 98.25 & 72.88 & 93.80 & 86.06 & 98.64 & 83.61 & 99.15 & 99.41 & 99.13 & 99.45 & 98.93 & 98.52 & 88.08 & \underline{99.48} & 97.43 & 83.49 & 98.69 & 93.37 & 71.18 & 95.70 \\
44 & 68.10 & 40.84 & \underline{75.97} & 71.73 & 46.69 & 51.40 & \textbf{78.13} & 66.60 & 45.89 & 48.82 & 32.70 & 30.91 & 42.49 & 29.97 & 36.15 & 33.40 & 32.83 & 32.67 & 52.51 & 33.36 & 37.26 & 42.43 & 30.45 & 52.02 & 42.93 & 49.00 \\
45 & 55.94 & 46.22 & 83.19 & 22.58 & 42.71 & 81.07 & 71.92 & 66.88 & 64.19 & 36.07 & 86.01 & 41.14 & 78.19 & 86.48 & 81.62 & \textbf{89.88} & 70.04 & 61.75 & 43.89 & \underline{87.09} & 74.13 & 72.52 & 86.09 & 41.00 & 54.60 & 41.13 \\
46 & 44.82 & 50.30 & \underline{54.01} & 48.86 & 46.70 & 46.34 & 50.33 & 50.83 & 50.06 & 41.60 & 50.69 & 49.38 & 49.23 & 48.40 & 47.94 & \textbf{54.78} & 47.86 & 47.34 & 51.21 & 52.34 & 48.39 & 52.35 & 50.39 & 53.45 & 47.46 & 46.61 \\
47 & 40.24 & 41.13 & 41.38 & 42.79 & 38.83 & 41.89 & 40.54 & 43.87 & 43.10 & 44.55 & 38.02 & 48.36 & 38.37 & 41.22 & 39.03 & 39.13 & 40.99 & 40.91 & \underline{50.08} & 36.89 & 44.00 & 38.08 & 46.61 & 42.83 & \textbf{59.32} & 43.89 \\
\midrule
Mean & 73.10 & 70.68 & 76.67 & 65.81 & 70.02 & \textbf{78.68} & \underline{77.28} & 65.13 & 75.95 & 61.74 & 73.77 & 72.14 & 76.39 & 74.01 & 74.04 & 74.47 & 70.97 & 71.19 & 61.98 & 74.49 & 74.44 & 65.81 & 66.02 & 69.30 & 48.73 & 62.82 \\
Median-Rank & 13.00 & 15.00 & 11.00 & 12.00 & 13.00 & \textbf{8.00} & 10.00 & 17.00 & 11.00 & 19.00 & 11.00 & 13.00 & \underline{9.00} & \underline{9.00} & \underline{9.00} & 10.00 & 16.00 & 16.00 & 20.00 & 11.00 & 10.00 & 19.00 & 17.00 & 16.00 & 23.00 & 19.00 \\
\bottomrule
\end{tabular}
}
\end{sidewaystable*}

\FloatBarrier

\section{Extended augmentation results}

The following section provides additional analysis of the data augmentation experiments introduced in Section~4.4 of the main paper under the TabEBM protocol. We report results obtained across ten random seeds to account for variability introduced during training and evaluation. The effectiveness of each augmentation strategy is evaluated using four downstream classifiers: K-Nearest Neighbors (KNN) (Table~\ref{tab:knn_aug}), Logistic Regression (LR) (Table~\ref{tab:lr_aug}), Multi-Layer Perceptron (MLP) (Table~\ref{tab:mlp_aug}), and Random Forest (RF) (Table~\ref{tab:rf_aug}). Performance is measured in terms of balanced accuracy. All classifiers use their scikit-learn implementations with default configurations. Finally, Table~\ref{tab:avg_aug} reports the average performance across all evaluated classifiers.

\begin{table*}[t]
\centering
\footnotesize
\caption{Balanced accuracy results of the K-Nearest Neighbors (KNN) classifier under different data augmentation strategies.}
\label{tab:knn_aug}
\begin{tabular}{lcccccccccc}
\toprule
dataset & n\_real & Baseline & SMOTE & TVAE & CTGAN & NFLOW & TabDDPM & ARF & TabEBM & ICED \\
\midrule
biodeg & 20 & \underline{69.35} & \textbf{71.14} & 60.91 & 62.33 & 63.24 & 67.92 & 64.01 & 69.12 & 69.06 \\
biodeg & 50 & \textbf{74.68} & 74.13 & 70.39 & 69.58 & 72.97 & \underline{74.66} & 70.22 & 72.00 & 72.06 \\
biodeg & 100 & \underline{78.85} & \textbf{78.92} & 74.13 & 77.40 & 78.82 & 78.83 & 76.53 & 77.11 & 77.15 \\
biodeg & 200 & \textbf{80.83} & 79.08 & 78.68 & 80.45 & 80.46 & \textbf{80.83} & 79.67 & 78.02 & 77.63 \\
biodeg & 500 & \textbf{82.71} & 81.56 & 82.45 & 82.68 & 82.62 & \textbf{82.71} & 82.18 & 80.77 & 81.87 \\
collins & 100 & 14.82 & 14.86 & 14.08 & 14.49 & 14.70 & 14.72 & 14.09 & \underline{15.91} & \textbf{16.06} \\
collins & 200 & \textbf{17.89} & 17.56 & 17.16 & 17.72 & 17.79 & \underline{17.87} & 17.82 & 17.28 & 17.27 \\
energy & 50 & 17.61 & 20.67 & 18.19 & 16.85 & 17.69 & 16.30 & 18.03 & \textbf{29.16} & \underline{28.46} \\
energy & 100 & 24.76 & 28.07 & 24.93 & 24.87 & 25.38 & 23.31 & 25.49 & \textbf{36.92} & \underline{35.62} \\
energy & 200 & 33.25 & 35.76 & 33.56 & 33.27 & 33.48 & 31.84 & 34.02 & \textbf{43.11} & \underline{41.02} \\
fourier & 20 & 33.30 & 34.93 & 23.92 & 31.40 & 29.98 & 33.30 & 34.02 & \textbf{46.16} & \underline{45.74} \\
fourier & 50 & 50.80 & 50.11 & 37.78 & 50.56 & 47.58 & 50.80 & 49.06 & \textbf{57.54} & \underline{57.46} \\
fourier & 100 & 60.50 & \textbf{68.32} & 49.74 & 60.46 & 57.50 & 60.50 & 59.02 & \underline{64.60} & \underline{64.60} \\
fourier & 200 & 68.00 & \textbf{71.54} & 60.76 & 68.16 & 66.92 & 68.00 & 66.72 & \underline{70.20} & 69.30 \\
fourier & 500 & \underline{74.96} & \textbf{75.62} & 73.72 & \underline{74.96} & 74.88 & \underline{74.96} & 74.34 & 74.24 & 73.94 \\
protein & 20 & 30.92 & 33.05 & 26.23 & 30.91 & 29.41 & 30.92 & 29.89 & \textbf{42.00} & \underline{41.98} \\
protein & 50 & 43.36 & 47.13 & 40.46 & 43.24 & 43.06 & 43.36 & 42.52 & \textbf{58.75} & \underline{58.54} \\
protein & 100 & 55.36 & \textbf{74.68} & 52.90 & 55.30 & 55.44 & 55.36 & 55.31 & \underline{73.55} & 73.36 \\
protein & 200 & 72.36 & \underline{84.20} & 71.12 & 72.36 & 72.36 & 72.36 & 72.82 & \textbf{84.63} & 83.11 \\
protein & 500 & 92.29 & \underline{95.07} & 92.32 & 92.35 & 92.35 & 92.29 & 92.39 & 94.85 & \textbf{95.08} \\
steel & 20 & 64.24 & \textbf{75.69} & 60.09 & 59.68 & 61.05 & 63.69 & 59.40 & \underline{74.51} & 74.12 \\
steel & 50 & 79.73 & \textbf{85.53} & 65.07 & 69.67 & 71.41 & 79.76 & 69.56 & 85.23 & \underline{85.26} \\
steel & 100 & 86.99 & \textbf{90.13} & 72.34 & 79.83 & 82.13 & 86.99 & 77.04 & \underline{89.29} & 88.69 \\
steel & 200 & 91.92 & \textbf{93.21} & 79.80 & 88.76 & 89.93 & 91.92 & 86.46 & 92.62 & \underline{92.71} \\
steel & 500 & 96.15 & \textbf{96.51} & 94.64 & 95.23 & 95.99 & 96.15 & 94.73 & 96.42 & \underline{96.49} \\
stock & 20 & 77.72 & \underline{84.15} & 76.85 & 73.90 & 77.95 & 78.81 & 79.53 & \textbf{84.20} & 83.57 \\
stock & 50 & 83.71 & 89.14 & 84.68 & 83.14 & 84.71 & 83.88 & 86.44 & \underline{89.52} & \textbf{89.65} \\
stock & 100 & 90.22 & \textbf{93.00} & 90.03 & 89.84 & 90.39 & 90.22 & 89.68 & \underline{92.93} & 92.82 \\
stock & 200 & 92.44 & 93.52 & 92.50 & 92.53 & 92.44 & 92.44 & 92.38 & \textbf{93.73} & \underline{93.70} \\
texture & 50 & 68.37 & 71.02 & 66.90 & 68.09 & 68.49 & 68.53 & 67.09 & \underline{78.61} & \textbf{79.45} \\
texture & 100 & 78.50 & \textbf{86.45} & 78.11 & 78.48 & 78.50 & 78.25 & 78.23 & \underline{84.92} & 84.88 \\
texture & 200 & 87.31 & \textbf{90.73} & 87.23 & 87.31 & 87.31 & 86.94 & 87.27 & 89.99 & \underline{90.01} \\
texture & 500 & 92.76 & 93.21 & 92.70 & 92.76 & 92.76 & 92.76 & 92.82 & \textbf{93.68} & \underline{93.40} \\
\midrule
Mean & - & 65.66 & 69.05 & 61.95 & 64.20 & 64.54 & 65.49 & 64.21 & \textbf{70.65} & \underline{70.43} \\
Median-Rank & - & 4.50 & \textbf{2.00} & 9.00 & 7.00 & 6.00 & 5.00 & 7.00 & \textbf{2.00} & \textbf{2.00} \\
\bottomrule
\end{tabular}
\end{table*}

\begin{table*}[t]
\centering
\footnotesize
\caption{Balanced accuracy results of the Logistic Regression (LR) classifier for different augmentation methods.}
\label{tab:lr_aug}
\begin{tabular}{lcccccccccc}
\toprule
dataset & n\_real & Baseline & SMOTE & TVAE & CTGAN & NFLOW & TabDDPM & ARF & TabEBM & ICED \\
\midrule
biodeg & 20 & \textbf{71.54} & \underline{70.78} & 60.90 & 59.63 & 52.48 & 55.37 & 68.08 & 69.90 & 70.05 \\
biodeg & 50 & \textbf{77.37} & \underline{76.62} & 70.26 & 62.47 & 57.86 & 58.44 & 72.13 & 76.07 & 75.74 \\
biodeg & 100 & \underline{80.36} & \textbf{80.66} & 76.27 & 72.38 & 64.26 & 68.16 & 76.59 & 79.86 & \underline{80.36} \\
biodeg & 200 & \textbf{82.13} & \underline{82.10} & 78.75 & 75.46 & 70.20 & 76.09 & 79.03 & 81.68 & 81.76 \\
biodeg & 500 & 83.65 & \underline{83.74} & 81.20 & 80.23 & 78.84 & 78.70 & 81.93 & \textbf{83.97} & 83.42 \\
collins & 100 & \textbf{15.82} & 14.43 & 14.94 & 14.34 & 13.51 & 11.43 & 15.51 & \underline{15.67} & 15.58 \\
collins & 200 & \textbf{19.56} & 18.29 & 19.04 & 18.49 & 17.76 & 16.41 & 18.46 & \underline{19.21} & 18.66 \\
energy & 50 & 22.20 & 19.71 & 20.00 & 18.09 & 13.96 & 14.38 & 20.77 & \textbf{26.21} & \underline{24.57} \\
energy & 100 & 26.04 & 25.06 & 25.10 & 24.43 & 22.84 & 15.23 & 24.88 & \textbf{31.99} & \underline{30.93} \\
energy & 200 & 32.15 & 31.03 & 32.38 & 31.32 & 29.71 & 19.48 & 31.51 & \textbf{37.08} & \underline{35.73} \\
fourier & 20 & \textbf{50.78} & 34.17 & 23.10 & 20.42 & 22.48 & 24.18 & 48.64 & 50.18 & \underline{50.20} \\
fourier & 50 & \textbf{63.10} & 44.96 & 34.42 & 35.52 & 30.90 & 34.46 & 54.44 & \underline{62.64} & 62.36 \\
fourier & 100 & \textbf{69.82} & 69.10 & 46.90 & 45.56 & 42.32 & 42.44 & 55.26 & \underline{69.48} & 68.98 \\
fourier & 200 & \textbf{74.12} & 73.18 & 58.54 & 56.72 & 55.94 & 52.00 & 60.82 & \underline{73.52} & 73.38 \\
fourier & 500 & \textbf{75.62} & \underline{75.02} & 69.96 & 68.12 & 68.70 & 66.52 & 70.58 & 74.92 & 74.80 \\
protein & 20 & 47.22 & 36.24 & 32.88 & 25.55 & 18.93 & 34.20 & 46.37 & \underline{47.86} & \textbf{47.88} \\
protein & 50 & 70.78 & 53.20 & 50.11 & 39.12 & 32.43 & 48.51 & 59.16 & \textbf{71.67} & \underline{71.39} \\
protein & 100 & 83.86 & 83.70 & 63.47 & 56.55 & 49.58 & 58.28 & 62.48 & \textbf{84.17} & \underline{84.09} \\
protein & 200 & \underline{93.77} & 93.36 & 79.71 & 72.62 & 70.10 & 72.03 & 78.03 & \textbf{94.17} & 93.45 \\
protein & 500 & \underline{98.30} & 98.17 & 90.80 & 88.55 & 87.00 & 79.61 & 90.08 & 98.28 & \textbf{98.32} \\
steel & 20 & 77.59 & 79.28 & 59.56 & 60.28 & 51.91 & 58.33 & 58.28 & \textbf{80.35} & \underline{80.22} \\
steel & 50 & 94.89 & 95.51 & 62.94 & 63.71 & 53.93 & 67.75 & 67.30 & \textbf{97.18} & \underline{97.10} \\
steel & 100 & 99.10 & 99.08 & 69.72 & 66.84 & 57.31 & 85.61 & 69.06 & \underline{99.28} & \textbf{99.42} \\
steel & 200 & 99.80 & 99.80 & 71.74 & 79.01 & 64.15 & 94.87 & 82.24 & \underline{99.81} & \textbf{99.84} \\
steel & 500 & \underline{99.84} & \underline{99.84} & 92.24 & 96.19 & 88.10 & 97.98 & 97.99 & 99.82 & \textbf{99.87} \\
stock & 20 & 78.01 & 79.78 & 75.66 & 66.04 & 75.64 & 59.76 & 70.77 & \underline{79.92} & \textbf{80.07} \\
stock & 50 & 81.12 & \underline{81.92} & 76.02 & 65.91 & 72.97 & 57.93 & 76.35 & \textbf{82.48} & 81.46 \\
stock & 100 & 81.75 & \textbf{83.50} & 78.11 & 75.01 & 75.15 & 62.86 & 77.94 & \underline{82.98} & 82.12 \\
stock & 200 & 82.67 & \textbf{84.11} & 77.64 & 77.84 & 75.36 & 69.91 & 78.88 & \underline{83.38} & 82.23 \\
texture & 50 & 89.94 & 60.57 & 60.61 & 56.37 & 36.12 & 39.00 & 50.64 & \textbf{90.79} & \underline{90.49} \\
texture & 100 & 93.99 & 93.59 & 76.00 & 73.86 & 60.10 & 62.41 & 68.39 & \textbf{94.39} & \underline{94.25} \\
texture & 200 & \underline{96.74} & 96.62 & 86.49 & 86.82 & 76.34 & 83.52 & 87.16 & \textbf{96.98} & 96.68 \\
texture & 500 & 98.02 & \underline{98.10} & 90.43 & 93.31 & 91.17 & 90.52 & 94.91 & \textbf{98.16} & 98.04 \\
\midrule
Mean & - & 73.08 & 70.16 & 60.78 & 58.39 & 53.88 & 56.25 & 63.47 & \textbf{73.76} & \underline{73.44} \\
Median-Rank & - & 3.00 & 4.00 & 6.00 & 7.00 & 9.00 & 8.00 & 5.00 & \textbf{2.00} & \underline{2.50} \\
\bottomrule
\end{tabular}
\end{table*}

\begin{table*}[t]
\centering
\footnotesize
\caption{Evaluation of data augmentation strategies using a Multi-Layer Perceptron (MLP) classifier. Results are reported in terms of balanced accuracy averaged over ten independent runs.}
\label{tab:mlp_aug}
\begin{tabular}{lcccccccccc}
\toprule
dataset & n\_real & Baseline & SMOTE & TVAE & CTGAN & NFLOW & TabDDPM & ARF & TabEBM & ICED \\
\midrule
biodeg & 20 & \textbf{72.22} & \underline{72.21} & 64.00 & 64.30 & 60.90 & 57.98 & 66.50 & 71.76 & 71.85 \\
biodeg & 50 & \textbf{76.48} & 75.67 & 71.81 & 70.36 & 72.31 & 71.60 & 73.07 & 75.99 & \underline{76.22} \\
biodeg & 100 & \textbf{80.63} & \underline{80.23} & 76.30 & 77.87 & 77.73 & 78.41 & 76.92 & 79.74 & 80.01 \\
biodeg & 200 & \underline{82.48} & 82.15 & 80.09 & 80.98 & 81.17 & \textbf{82.70} & 79.50 & 82.15 & 82.24 \\
biodeg & 500 & 84.25 & 84.28 & 83.16 & 83.67 & 83.74 & 84.57 & 82.58 & \textbf{84.86} & \underline{84.85} \\
collins & 100 & 16.54 & 15.76 & 15.33 & 15.81 & 14.63 & 14.75 & 15.95 & \underline{16.69} & \textbf{17.16} \\
collins & 200 & 19.48 & 19.33 & 18.90 & 19.62 & 19.05 & 19.25 & 19.19 & \underline{19.63} & \textbf{19.65} \\
energy & 50 & \underline{27.50} & 25.90 & 24.78 & 24.54 & 25.08 & 26.68 & 24.79 & \textbf{28.16} & 27.28 \\
energy & 100 & 37.40 & 37.33 & 36.66 & 36.71 & 37.21 & \underline{38.42} & 35.93 & \textbf{39.04} & 37.33 \\
energy & 200 & 49.03 & 48.70 & 47.42 & 48.75 & 48.11 & \textbf{51.15} & 46.99 & \underline{49.49} & 48.97 \\
fourier & 20 & 44.76 & 34.09 & 23.04 & 26.24 & 26.22 & 28.46 & 44.36 & \textbf{45.20} & \underline{45.08} \\
fourier & 50 & 57.58 & 47.99 & 37.78 & 43.08 & 37.36 & 47.70 & 52.36 & \textbf{58.92} & \underline{58.72} \\
fourier & 100 & 65.80 & 65.80 & 48.74 & 53.68 & 49.48 & 60.18 & 61.36 & \textbf{66.66} & \underline{66.52} \\
fourier & 200 & 71.68 & 72.38 & 59.70 & 64.78 & 60.56 & 69.82 & 67.28 & \underline{72.72} & \textbf{73.04} \\
fourier & 500 & \underline{77.32} & 77.00 & 70.80 & 72.58 & 71.62 & 76.50 & 72.48 & \textbf{77.38} & 77.14 \\
protein & 20 & \underline{44.40} & 38.91 & 34.28 & 35.45 & 32.17 & 37.98 & 43.48 & \textbf{44.62} & 44.38 \\
protein & 50 & 67.12 & 59.74 & 50.66 & 55.05 & 54.63 & 61.80 & 60.89 & \textbf{67.86} & \underline{67.30} \\
protein & 100 & 81.00 & 81.16 & 65.05 & 73.30 & 72.80 & 77.85 & 72.28 & \underline{81.44} & \textbf{81.67} \\
protein & 200 & \underline{93.27} & 92.90 & 83.70 & 88.99 & 88.61 & 91.80 & 87.08 & \textbf{93.70} & 93.25 \\
protein & 500 & 99.08 & 98.87 & 96.66 & 97.78 & 97.89 & 98.75 & 97.69 & \underline{99.10} & \textbf{99.22} \\
steel & 20 & 73.94 & \textbf{75.18} & 59.76 & 65.29 & 59.50 & 63.93 & 59.66 & 74.78 & \underline{74.85} \\
steel & 50 & 87.23 & 87.29 & 69.90 & 75.04 & 71.90 & 82.51 & 66.18 & \textbf{89.41} & \underline{89.35} \\
steel & 100 & 95.70 & 96.04 & 74.70 & 82.68 & 81.47 & 95.21 & 73.15 & \underline{96.73} & \textbf{96.90} \\
steel & 200 & 98.79 & 99.00 & 82.10 & 91.35 & 88.89 & 98.32 & 83.74 & \underline{99.04} & \textbf{99.24} \\
steel & 500 & 99.54 & \underline{99.56} & 94.41 & 97.95 & 97.19 & 99.25 & 93.60 & \textbf{99.59} & \underline{99.56} \\
stock & 20 & 83.70 & 82.60 & 76.88 & 72.31 & 76.50 & 79.90 & 78.27 & \textbf{84.17} & \textbf{84.17} \\
stock & 50 & \textbf{90.20} & 89.40 & 86.75 & 85.43 & 86.25 & 87.14 & 86.28 & 89.97 & \underline{89.98} \\
stock & 100 & 93.64 & 93.29 & 91.88 & 91.96 & 90.84 & 92.04 & 91.02 & \underline{93.65} & \textbf{93.73} \\
stock & 200 & 94.19 & \textbf{94.42} & 92.88 & 93.29 & 93.61 & 92.74 & 92.64 & \underline{94.26} & 94.14 \\
texture & 50 & \underline{88.27} & 82.26 & 75.17 & 81.61 & 81.64 & 84.90 & 76.39 & \textbf{88.43} & 87.69 \\
texture & 100 & \underline{93.21} & 92.63 & 86.47 & 90.12 & 89.27 & 92.17 & 85.86 & \textbf{93.47} & 92.87 \\
texture & 200 & \underline{96.27} & 96.01 & 93.55 & 94.97 & 94.79 & 95.59 & 91.95 & \textbf{96.43} & 96.09 \\
texture & 500 & \textbf{98.18} & 97.88 & 97.40 & 97.22 & 97.30 & 97.42 & 96.62 & \underline{98.04} & 97.94 \\
\midrule
Mean & - & 73.97 & 72.60 & 65.78 & 68.27 & 67.29 & 70.83 & 68.36 & \textbf{74.34} & \underline{74.19} \\
Median-Rank & - & 3.00 & 4.00 & 8.00 & 6.00 & 7.00 & 5.00 & 8.00 & \textbf{2.00} & \textbf{2.00} \\
\bottomrule
\end{tabular}
\end{table*}

\begin{table*}[t]
\centering
\footnotesize
\caption{Performance comparison of augmentation approaches using a Random Forest (RF) classifier. The reported balanced accuracy values are averaged over ten random seeds.}
\label{tab:rf_aug}
\begin{tabular}{lcccccccccc}
\toprule
dataset & n\_real & Baseline & SMOTE & TVAE & CTGAN & NFLOW & TabDDPM & ARF & TabEBM & ICED \\
\midrule
biodeg & 20 & 70.15 & 70.76 & 62.51 & 62.10 & 59.71 & 61.70 & 65.02 & \underline{71.03} & \textbf{71.17} \\
biodeg & 50 & 72.12 & 72.92 & 69.80 & 66.05 & 65.02 & 69.87 & 69.09 & \underline{73.61} & \textbf{73.86} \\
biodeg & 100 & 76.23 & 77.61 & 73.81 & 74.01 & 72.46 & 74.96 & 74.34 & \underline{77.94} & \textbf{78.09} \\
biodeg & 200 & 79.55 & 79.95 & 78.20 & 77.27 & 78.52 & 79.41 & 78.53 & \textbf{80.52} & \underline{80.25} \\
biodeg & 500 & 82.50 & 82.51 & 81.85 & 82.18 & 82.25 & 82.18 & \underline{82.62} & \underline{82.62} & \textbf{82.79} \\
collins & 100 & 14.69 & 14.72 & 14.93 & 14.54 & 13.53 & 14.15 & 14.90 & \textbf{15.48} & \textbf{15.48} \\
collins & 200 & 17.21 & 17.50 & 16.82 & 17.36 & 17.59 & 17.63 & 17.51 & \underline{17.64} & \textbf{17.95} \\
energy & 50 & \textbf{39.41} & 37.26 & 37.89 & 37.92 & 37.96 & 32.30 & 38.21 & \underline{38.59} & 37.94 \\
energy & 100 & 50.32 & 50.05 & \underline{50.80} & 50.36 & 49.67 & 47.42 & 50.52 & \textbf{51.24} & 50.34 \\
energy & 200 & 59.56 & 59.57 & 59.74 & 60.04 & 59.76 & 57.06 & 59.74 & \underline{60.28} & \textbf{60.35} \\
fourier & 20 & 46.54 & 35.47 & 26.00 & 23.68 & 20.76 & 29.00 & 44.06 & \underline{51.76} & \textbf{53.02} \\
fourier & 50 & 65.98 & 54.78 & 46.54 & 47.60 & 33.76 & 59.14 & 58.66 & \textbf{69.50} & \underline{68.26} \\
fourier & 100 & 72.80 & 73.92 & 61.52 & 63.68 & 52.00 & 72.96 & 67.42 & \textbf{74.70} & \underline{74.20} \\
fourier & 200 & 76.38 & 77.04 & 71.42 & 73.64 & 68.66 & 76.30 & 73.74 & \underline{77.08} & \textbf{77.70} \\
fourier & 500 & \textbf{79.94} & 79.68 & 77.78 & 78.82 & 77.58 & 79.72 & 78.30 & 79.64 & \underline{79.82} \\
protein & 20 & 43.76 & 37.64 & 33.11 & 31.78 & 26.58 & 36.42 & 41.72 & \textbf{47.06} & \underline{46.77} \\
protein & 50 & 62.23 & 56.71 & 54.86 & 54.08 & 44.19 & 58.13 & 54.51 & \textbf{65.93} & \underline{65.24} \\
protein & 100 & 75.52 & 77.57 & 69.40 & 72.45 & 65.46 & 73.51 & 67.55 & \textbf{78.07} & \underline{77.83} \\
protein & 200 & 87.39 & 88.34 & 84.88 & 86.98 & 84.65 & 85.62 & 84.26 & \textbf{89.63} & \underline{88.77} \\
protein & 500 & 97.30 & \underline{97.37} & 96.62 & 96.89 & 96.86 & 96.65 & 96.39 & 97.26 & \textbf{97.49} \\
steel & 20 & 59.41 & \textbf{65.19} & 56.77 & 54.77 & 53.96 & 55.46 & 56.03 & \underline{62.68} & 62.41 \\
steel & 50 & 65.00 & \textbf{74.75} & 58.85 & 56.48 & 54.05 & 61.44 & 58.31 & \underline{73.68} & 73.28 \\
steel & 100 & 73.19 & 82.84 & 62.45 & 63.01 & 57.73 & 69.91 & 63.37 & \underline{84.87} & \textbf{86.31} \\
steel & 200 & 85.58 & 88.23 & 69.85 & 72.43 & 64.57 & 81.75 & 70.27 & \underline{92.08} & \textbf{93.63} \\
steel & 500 & 96.68 & 97.00 & 88.37 & 90.38 & 86.40 & 95.43 & 89.37 & \textbf{98.82} & \underline{98.57} \\
stock & 20 & 82.95 & 81.18 & 79.14 & 74.89 & 80.82 & 81.56 & 80.80 & \textbf{85.11} & \underline{84.40} \\
stock & 50 & 89.61 & \textbf{90.33} & 88.46 & 85.64 & 87.68 & 87.51 & 87.78 & \underline{90.02} & 89.60 \\
stock & 100 & \textbf{93.45} & 93.09 & 93.07 & 92.26 & 91.25 & 92.54 & 91.84 & \underline{93.16} & 93.10 \\
stock & 200 & 94.37 & \textbf{94.45} & 94.18 & 93.96 & 94.07 & 94.18 & 93.75 & 94.21 & \textbf{94.45} \\
texture & 50 & 75.44 & 72.37 & 64.66 & 74.08 & 73.11 & 73.50 & 67.02 & \underline{77.04} & \textbf{77.19} \\
texture & 100 & 82.42 & 82.90 & 75.04 & 82.60 & 81.62 & 81.28 & 77.24 & \underline{83.68} & \textbf{84.09} \\
texture & 200 & 88.01 & 88.82 & 86.80 & 88.39 & 88.65 & 87.92 & 88.14 & \underline{89.76} & \textbf{90.33} \\
texture & 500 & 93.19 & 93.06 & 92.40 & 93.13 & 92.98 & \underline{93.43} & 93.16 & \textbf{93.84} & 93.38 \\
\midrule
Mean & - & 71.18 & 71.08 & 66.02 & 66.47 & 64.06 & 68.49 & 67.70 & \textbf{73.29} & \underline{73.27} \\
Median-Rank & - & 4.00 & 3.00 & 8.00 & 7.00 & 9.00 & 5.00 & 6.00 & \textbf{2.00} & \textbf{2.00} \\
\bottomrule
\end{tabular}
\end{table*}

\begin{table*}[t]
\centering
\footnotesize
\caption{Average balanced accuracy across four downstream classifiers (KNN, LR, MLP, and RF) for different data augmentation techniques. The table summarizes the overall effectiveness of each augmentation strategy across classifiers and datasets.}
\label{tab:avg_aug}
\begin{tabular}{lcccccccccc}
\toprule
dataset & n\_real & Baseline & SMOTE & TVAE & CTGAN & NFLOW & TabDDPM & ARF & TabEBM & ICED \\
\midrule
biodeg & 20 & \underline{70.82} & \textbf{71.22} & 62.08 & 62.09 & 59.08 & 60.74 & 65.90 & 70.45 & 70.53 \\
biodeg & 50 & \textbf{75.16} & \underline{74.83} & 70.56 & 67.11 & 67.04 & 68.64 & 71.13 & 74.42 & 74.47 \\
biodeg & 100 & \underline{79.02} & \textbf{79.35} & 75.13 & 75.41 & 73.31 & 75.09 & 76.10 & 78.66 & 78.90 \\
biodeg & 200 & \textbf{81.25} & \underline{80.82} & 78.93 & 78.54 & 77.59 & 79.76 & 79.18 & 80.59 & 80.47 \\
biodeg & 500 & \textbf{83.28} & 83.02 & 82.16 & 82.19 & 81.86 & 82.04 & 82.33 & 83.06 & \underline{83.23} \\
collins & 100 & 15.47 & 14.94 & 14.82 & 14.79 & 14.09 & 13.76 & 15.11 & \underline{15.94} & \textbf{16.07} \\
collins & 200 & \textbf{18.54} & 18.17 & 17.98 & 18.30 & 18.05 & 17.79 & 18.24 & \underline{18.44} & 18.38 \\
energy & 50 & 26.68 & 25.89 & 25.22 & 24.35 & 23.67 & 22.42 & 25.45 & \textbf{30.53} & \underline{29.57} \\
energy & 100 & 34.63 & 35.13 & 34.37 & 34.09 & 33.77 & 31.09 & 34.20 & \textbf{39.80} & \underline{38.55} \\
energy & 200 & 43.50 & 43.76 & 43.28 & 43.35 & 42.76 & 39.88 & 43.06 & \textbf{47.49} & \underline{46.52} \\
fourier & 20 & 43.84 & 34.66 & 24.02 & 25.44 & 24.86 & 28.74 & 42.77 & \underline{48.32} & \textbf{48.51} \\
fourier & 50 & 59.36 & 49.46 & 39.13 & 44.19 & 37.40 & 48.02 & 53.63 & \textbf{62.15} & \underline{61.70} \\
fourier & 100 & 67.23 & \textbf{69.28} & 51.72 & 55.84 & 50.32 & 59.02 & 60.76 & \underline{68.86} & 68.58 \\
fourier & 200 & 72.54 & \textbf{73.54} & 62.60 & 65.82 & 63.02 & 66.53 & 67.14 & \underline{73.38} & 73.35 \\
fourier & 500 & \textbf{76.96} & \underline{76.83} & 73.06 & 73.62 & 73.19 & 74.42 & 73.92 & 76.54 & 76.42 \\
protein & 20 & 41.57 & 36.46 & 31.63 & 30.92 & 26.77 & 34.88 & 40.36 & \textbf{45.38} & \underline{45.25} \\
protein & 50 & 60.87 & 54.19 & 49.02 & 47.87 & 43.58 & 52.95 & 54.27 & \textbf{66.05} & \underline{65.62} \\
protein & 100 & 73.93 & \underline{79.27} & 62.71 & 64.40 & 60.82 & 66.25 & 64.40 & \textbf{79.31} & 79.24 \\
protein & 200 & 86.70 & \underline{89.70} & 79.85 & 80.24 & 78.93 & 80.45 & 80.55 & \textbf{90.53} & 89.65 \\
protein & 500 & 96.74 & \underline{97.37} & 94.10 & 93.89 & 93.52 & 91.82 & 94.14 & \underline{97.37} & \textbf{97.53} \\
steel & 20 & 68.79 & \textbf{73.83} & 59.04 & 60.00 & 56.61 & 60.35 & 58.34 & \underline{73.08} & 72.90 \\
steel & 50 & 81.71 & 85.77 & 64.19 & 66.22 & 62.83 & 72.87 & 65.34 & \textbf{86.38} & \underline{86.24} \\
steel & 100 & 88.74 & 92.02 & 69.80 & 73.09 & 69.66 & 84.43 & 70.66 & \underline{92.54} & \textbf{92.83} \\
steel & 200 & 94.03 & 95.06 & 75.87 & 82.89 & 76.88 & 91.72 & 80.68 & \underline{95.89} & \textbf{96.36} \\
steel & 500 & 98.05 & 98.23 & 92.41 & 94.94 & 91.92 & 97.20 & 93.92 & \textbf{98.67} & \underline{98.62} \\
stock & 20 & 80.60 & 81.93 & 77.13 & 71.79 & 77.73 & 75.01 & 77.34 & \textbf{83.35} & \underline{83.05} \\
stock & 50 & 86.16 & \underline{87.70} & 83.98 & 80.03 & 82.90 & 79.11 & 84.21 & \textbf{88.00} & 87.67 \\
stock & 100 & 89.76 & \textbf{90.72} & 88.27 & 87.26 & 86.91 & 84.41 & 87.62 & \underline{90.68} & 90.44 \\
stock & 200 & 90.92 & \textbf{91.62} & 89.30 & 89.40 & 88.87 & 87.32 & 89.41 & \underline{91.39} & 91.13 \\
texture & 50 & 80.51 & 71.56 & 66.83 & 70.04 & 64.84 & 66.48 & 65.29 & \textbf{83.72} & \underline{83.70} \\
texture & 100 & 87.03 & 88.89 & 78.90 & 81.26 & 77.37 & 78.53 & 77.43 & \textbf{89.12} & \underline{89.02} \\
texture & 200 & 92.08 & 93.05 & 88.52 & 89.37 & 86.78 & 88.49 & 88.63 & \textbf{93.29} & \underline{93.28} \\
texture & 500 & 95.54 & 95.56 & 93.23 & 94.11 & 93.55 & 93.53 & 94.38 & \textbf{95.93} & \underline{95.69} \\
\midrule
Mean & - & 70.97 & 70.72 & 63.63 & 64.33 & 62.44 & 65.26 & 65.94 & \textbf{73.01} & \underline{72.83} \\
Median-Rank & - & 4.00 & 3.00 & 7.00 & 7.00 & 9.00 & 7.00 & 5.00 & \textbf{2.00} & \textbf{2.00} \\
\bottomrule
\end{tabular}
\end{table*}

\begin{table*}[!httb]
\centering
\footnotesize
\caption{Kendall's $\tau$ results for the prior ablation study on held-out synthetic datasets.}
\label{tab:prior_ablation_kendall}
\begin{tabular}{lcc|cccccc}
\toprule
\multicolumn{3}{c|}{Evaluation datasets} & \multicolumn{6}{c}{Prior} \\
Base class & Cat. & Aug. & w/o h-tail & w/o Gauss & w/o flow & w/o cat & w/o augm & full prior \\
\midrule
Gaussian mix. & \cmark & \xmark & \textbf{0.5787} & 0.5544 & 0.5717 & 0.5124 & \underline{0.5739} & 0.5648 \\
Heavy-tailed mix. & \cmark & \xmark & 0.4631 & \underline{0.5097} & 0.5077 & 0.4742 & \textbf{0.5185} & 0.4925 \\
% Random mix. (cat) & 0.7252 & 0.7327 & \textbf{0.7360} & 0.7237 & \underline{0.7345} & 0.7306 \\
% Flow-transformed (cont) & 0.7631 & 0.7708 & 0.7624 & \underline{0.7718} & \textbf{0.7752} & 0.7671 \\
Mix. with flows & \cmark & \xmark & 0.5020 & \underline{0.5087} & 0.4973 & 0.4742 & \textbf{0.5218} & 0.5082 \\
 All datasets & \xmark & \cmark & 0.4708 & 0.4756 & 0.4700 & \textbf{0.4830} & 0.4710 & \underline{0.4761} \\
All datasets & \cmark & \cmark & \underline{0.4397} & 0.4386 & 0.4356 & 0.4019 & 0.4327 & \textbf{0.4418} \\
% \midrule
% Mean & 0.7449 & \underline{0.7494} & 0.7478 & 0.7383 & \textbf{0.7505} & 0.7485 \\
\bottomrule
\end{tabular}
\end{table*}

\begin{table*}[t]
\centering
\setlength{\tabcolsep}{4pt}
\footnotesize
\caption{Influence of individual loss components on density estimation performance measured by Kendall's $\tau$. The full objective combines the ranking and MSE losses, while the remaining variants remove one of these terms.}
\label{tab:loss_ablation_kendall}
\begin{tabular}{lcc|ccc}
\toprule
\multicolumn{3}{c|}{Evaluation datasets} & \multicolumn{3}{c}{Loss} \\
Base class & Cat. & Aug. & w/o rank & w/o MSE & full loss \\
\midrule
Gaussian mix.  & \cmark & \xmark & \underline{0.5531} & 0.5083 & \textbf{0.5648} \\
Heavy-tailed mix. & \cmark & \xmark & \underline{0.4878} & 0.4605 & \textbf{0.4925} \\
Mix. with flows & \cmark & \xmark & \underline{0.4998} & 0.4680 & \textbf{0.5082} \\
All datasets & \xmark & \cmark & \underline{0.4697} & 0.4199 & \textbf{0.4761} \\
All datasets & \cmark & \cmark & \underline{0.4314} & 0.3940 & \textbf{0.4418} \\
\bottomrule
\end{tabular}
\end{table*}

\FloatBarrier
\section{Additional results for ablation studies}
\label{app:augmentation_exp}

Similar to Appendix~\ref{app:density-kendall}, this section reports additional
Kendall's $\tau$ results for the ablation experiments. We evaluate the impact of
different prior choices (Table~\ref{tab:prior_ablation_kendall}) and loss
function variants (Table~\ref{tab:loss_ablation_kendall}) using the same
experimental setting as in the main paper.

\FloatBarrier

\section{Computational Infrastructure}

All experiments were conducted on a GPU workstation equipped with an NVIDIA DGX A100 system with an NVIDIA A100 GPU (80 GB VRAM) and an AMD Ryzen Threadripper PRO 5975WX CPU with 256 GB RAM. The experiments were run under a Linux operating system. The implementations were developed in Python using the PyTorch deep learning framework together with standard scientific computing and machine learning libraries, including NumPy, SciPy, pandas, and scikit-learn. The specific library versions and software dependencies are provided in the code repository available in the code appendix.

\end{document}